\documentclass[twoside,11pt,preprint]{article}
\usepackage[T1]{fontenc}
\usepackage{jmlr2e}

\usepackage{float}
\usepackage{booktabs}
\usepackage{multirow}
\usepackage{tablefootnote}
\usepackage{mathtools} 
\usepackage{cleveref}
\usepackage{tikz}
\usetikzlibrary{arrows.meta,matrix,automata,shapes,positioning}
\tikzstyle{elbl} = [fill=white,inner sep=1pt,font=\footnotesize,>=stealth]

\usepackage{siunitx} 
\newcommand*{\Alpha}{\texttt{Alph}}
\newcommand*{\Tier}{\texttt{Tier}}
\newcommand*{\Class}{\texttt{Class}}
\newcommand*{\kay}{\texttt{k}}
\newcommand*{\jay}{\texttt{j}}
\newcommand*{\eye}{\texttt{i}}
\newcommand*{\TrainSize}{\texttt{TrainSize}}
\newcommand*{\TestType}{\texttt{TestType}}
\newcommand*{\NNType}{\texttt{NNType}}
\newcommand*{\scr}{\lhd}

\newcommand{\centerdash}{\multicolumn{1}{c}{\textendash}}

\usepackage{lastpage}
\jmlrheading{25}{2024}{1-\pageref{LastPage}}{4/23; Revised 7/24}{8/24}{23-0518}{Sam van der
  Poel, Dakotah Lambert, Kalina Kostyszyn, Tiantian Gao, Rahul Verma,
  Derek Andersen, Joanne Chau, Emily Peterson, Cody St. Clair, Paul
  Fodor, Chihiro Shibata, and Jeffrey Heinz}

\ShortHeadings{MLRegTest}{van der Poel et al.}

\firstpageno{1}

\begin{document}

\title{MLRegTest: A Benchmark for the Machine Learning of Regular Languages}

\author{\name Sam van der Poel \email samvanderpoel@gatech.edu\\
  \addr School of Mathematics\\
  Georgia Institute of Technology
  \AND
  \name Dakotah Lambert \email dakotahlambert@acm.org\\
  \addr Department of Computer Science \\
  Haverford College
  \AND
  \name Kalina Kostyszyn \email kalina.kostyszyn@stonybrook.edu\\
  \addr Department of Linguistics \&\\
  Institute of Advanced Computational Science\\
  Stony Brook University
  \AND
  \name Tiantian Gao \email tiagao@cs.stonybrook.edu\\
  \name Rahul Verma \email rxverma1@gmail.com\\ 
  \addr Department of Computer Science\\ Stony Brook University
  \AND
  \name Derek Andersen\email derek.andersen@alumni.stonybrook.edu\\
  \name Joanne Chau \email choryan.chau@alumni.stonybrook.edu\\
  \name Emily Peterson \email emily.peterson@alumni.stonybrook.edu\\ 
  \name Cody St. Clair \email cody.stclair@alumni.stonybrook.edu\\
  \addr Department of Linguistics\\ Stony Brook University
  \AND
  \name Paul Fodor \email pfodor@cs.stonybrook.edu\\
  \addr Department of Computer Science\\ Stony Brook University
  \AND
  \name Chihiro Shibata \email chihiro@hosei.ac.jp\\
  \addr Department of Advanced Sciences\\
  Graduate School of Science and Engineering\\
  Hosei University
  \AND
  \name Jeffrey Heinz \email jeffrey.heinz@stonybrook.edu\\
  \addr Department of Linguistics \&\\
  Institute of Advanced Computational Science\\
  Stony Brook University
}

\editor{Alexander Clark}
\maketitle

\newpage

\begin{abstract}
  Synthetic datasets constructed from formal languages allow
  fine-grained examination of the learning and generalization
  capabilities of machine learning systems for sequence
  classification. This article presents a new benchmark for machine
  learning systems on sequence classification called MLRegTest, which
  contains training, development, and test sets from 1,800 regular
  languages.

  Different kinds of formal languages represent different kinds of
  long-distance dependencies, and correctly identifying long-distance
  dependencies in sequences is a known challenge for ML systems to
  generalize successfully.  MLRegTest organizes its languages
  according to their logical complexity (monadic second order, first
  order, propositional, or restricted propositional) and the
  kind of logical literals (string, tier-string, subsequence, or
  combinations thereof). The logical complexity and choice of literal
  provides a systematic way to understand different kinds of
  long-distance dependencies in regular languages, and therefore to
  understand the capacities of different ML systems to learn such
  long-distance dependencies.
  
  Finally, the performance of different neural networks (simple RNN,
  LSTM, GRU, transformer) on MLRegTest is examined. The main
  conclusion is that performance depends significantly on the
  kind of test set, the class of language, and the neural network
  architecture.
\end{abstract}


\section{Introduction} 
This article presents a new benchmark for the machine learning (ML) of
regular languages called MLRegTest.\footnote{MLRegTest is publicly
  available with Dryad \url{https://doi.org/10.5061/dryad.dncjsxm4h}
  under the license \emph{CC0 1.0 Universal (CC0 1.0) Public Domain
    Dedication}
  (\url{https://creativecommons.org/publicdomain/zero/1.0/}). Software
  used to create and run the experiments in this paper are available
  in a Github repository at
  \url{https://github.com/heinz-jeffrey/subregular-learning} under a
  \emph{Creative Commons Attribution 4.0 International License}
  (\url{https://creativecommons.org/licenses/by/4.0/}).} Regular
languages are formal languages, which are sets of sequences definable
with certain kinds of formal grammars, including regular expressions,
finite-state acceptors, and monadic second order logic with either the
successor or precedence relation in the model signature for words
\citep{Kleene1956,ScottRabin1959,Buchi1960}.

One way to investigate the capacities of ML systems is to examine
their performance on data generated from processes which are known. If
ML systems perform well on such data, it builds confidence when the
same ML systems are applied to learning patterns from data generated
from unknown sources.  In this way, MLRegTest allows one to better
understand the learning capabilities and limitations of practical ML
systems on learning patterns over sequences.  In addition, this
benchmark was specifically designed to help identify those factors,
specifically the kinds of long-distance dependencies, that can make it
difficult for ML systems to successfully learn to classify
sequences. MLRegTest contains 1,800 languages from 16 distinct
subregular classes whose formal properties are well-understood. It is
the most comprehensive suite of regular languages we are aware
of. Finally, experimental results on the benchmark can be aggregated
to form a complete block design, which facilitates statistical
analysis of the results.

For each language, the benchmark includes three nested training sizes
with equal numbers of positive and negative examples, three nested
development sizes with equal numbers of positive and negative
examples, and three nested sizes of four distinct test sets with equal
numbers of positive and negative examples. The four test sets
manipulate two ways in which testing can be difficult: (1) the test
strings can either be at most as long as the longest training strings or they
can be longer, and (2) the test strings can either be randomly generated or
designed to occur in pairs of strings \(x\) and \(y\) such that
\(x\in L\), \(y\not\in L\) and the string edit distance of \(x\) and \(y\)
equals 1. We refer to such pairs of strings as forming the `border' of
the language. 

Another aspect of MLRegTest's design was its attention to the role of
long-distance dependencies in sequence classification. Long-distance
dependencies are widely recognized as a key challenge to generalizing
successfully.  \citet{BENGIO1994} define long-term dependencies this
way: ``A task displays long-term dependencies if prediction of the
desired output at time \(t\) depends on input presented at an earlier
time \(\tau\ll t\).''  Many examples of such long-term dependencies
abound in nature and engineering. For example, generative linguists,
beginning with \citet{chomsky56,chomsky57}, have studied the
grammatical basis of long-term dependencies in natural languages and
have raised the question of how such dependencies are learned
\citep{chomsky65}. However, there are many different ways in which a long-term
dependency can manifest itself, and we should be interested in
classifying long-term dependencies to the same degree as we are interested in
classifying types of non-linear, numerical functions.

Formal languages provide a way to achieve such a classification, and
the 16 classes used in this article are characterized by the kinds of
long-term dependencies required to successfully distinguish strings.
MLRegTest organizes its languages along two dimensions. One is
according to their logical complexity. Can the formal language be
expressed with a sentence of monadic second order logic, first order
logic, or a propositional logic, potentially with restrictions? The
other is according to the kind of logical literal. Are the primitives
in the logical language based on the notion of a string (successive
symbols), a tier-string (successive salient symbols after deleting all
symbols not in the so-called tier), a subsequence (not-necessarily
adjacent symbols in order), or combinations thereof? The logical
complexity and choice of literal provides a systematic way to
understand different kinds of long-distance dependencies in regular
languages.  In this way, we can study precisely the challenges certain
kinds of long-distance dependencies, in terms of their logical
complexity, bring to the learning of sequential classifiers.

We examine one such experimental design to broadly consider the
question of where the difficulties lie for neural networks (NNs)
learning to classify sequences drawn from regular languages from
positive and negative examples.  While we acknowledge that there may
exist some ML system we did not consider whose performance erases the
distinctions we find, our main objective was the development of the
benchmark. Our investigation suggests that it will be an important
milestone if other researchers are able to find an ML system that
succeeds across the board on MLRegTest.

From our experiments, we were able to draw two main conclusions.
First, neural networks generally perform worse on the test sets which
examine the border of the language. Consequently, performance on
randomly generated test data can mislead researchers into believing
correct generalization has been obtained, and stricter testing can
reveal it has not. This is not the first time such an observation has
been made \citep[and others]{pmlr-v80-weiss18a}, but the degree to
which it is observed here is striking.


Second, the formal properties of the languages are important in
determining its learning difficulty. It is not solely the size of the
grammatical representation that matters. This conclusion follows from
two findings. First we find that neither the size of the minimal
deterministic finite-state acceptor nor the size of its syntactic
monoid, which are two mathematically natural ways to measure the size
of a finite-state machine (see \S\ref{sec:lgs}), correlate especially
well with NN performance. We also find that, across the board, neural
networks have difficulty learning periodic regular languages; i.e
those that require monadic second order logic. Also, the neural
networks generally performed better on classifying strings on
languages which are defined logically with the successor relation
(which picks out adjacent elements in a string) as opposed to
languages which are defined logically with an order relation that
picks out non-adjacent elements in a string (the precedence or
tier-successor relations, see \S\ref{sec:lgs}). While there could be
other measures of grammar size that do correlate with learning
difficulty, it remains an open question what those grammatical
representations would be.

\section{Background and Related Work}

There is much precedent in exploring the use of formal languages to
investigate the learning capabilities of machine learning systems, and
neural networks in particular. Indeed this history goes right back to
the foundational chapters in computer science. For example, the
introduction of regular expressions into computer science
\citep{Kleene1956} was primarily motivated to understand the nerve
nets of \citep{mcculloch1943logical}. This kind of theoretical work
which establishes equivalencies and relationships between neural
network architectures and formal grammars continues to the present day
\citep{LiPrecupRabusseau2024}.

The reasons for making formal languages the targets of learning are as
valid today as they were decades ago.  First, the grammars generating
the formal languages are known and understood. Therefore training and
test data can be generated as desired to run controlled experiments to
see whether particular generalizations are reliably acquired under
particular training regimes.

Regular languages have often been used to benchmark ML
systems. \citet{Casey1996} and \citet{SmithA.W.1989} studied how well
first-order RNNs can learn to predict the next symbol of a string
using regular languages based on the Reber grammar \citep{REBER1967}.
\citet{Pollack1991}, \citet{Watrous1992}, and \citet{Giles1992} studied how
well second-order RNNs could learn to discriminate strings on the
Tomita regular languages \citep{Tomita1982}. Over time, the Tomita
languages have become a de facto benchmark for learning regular
languages \citep{Zeng+1994,pmlr-v80-weiss18a}.

Later research also targeted nonregular languages
\citep{Schmidhuber2002,Chalup2003955,PerezOrtiz2003241}. Readers are
encouraged to read \citet[sec.~5.13]{Schmidhuber2015}, which provides
an extensive review of this literature up to 2015, with extensive
focus on neural network ML architectures. More recent contributions in
this area include \citet{sennhauser-berwick-2018-evaluating,
  skachkova-etal-2018-closing, bhattamishra-etal-2020-ability,
  ebrahimi-etal-2020-self, Deletang+2022} and \citet{Merrill2023}.

There are some key differences between the present paper and past
research. First, the regular languages chosen here are known to have
certain properties.  The Reber grammars and Tomita languages were not
understood in terms of their abstract properties or pattern
complexity.  While it was recognized some encoded a long-distance
dependency and some did not, there was little recognition of the
computational nature of these formal languages beyond that. In
contrast, the formal languages in this paper are much better
understood.  While \emph{subregular} distinctions had already been
studied by the time of that research \citep{McNaughtonPapert1971}, it
went unrecognized how that branch of computer science could inform
machine learning. Since then, there has been much work on clarifying
particular subregular classes of languages in terms of their logical
complexity as well as their significance for cognition
\citep{RogersPullum2011, Rogers-HeinzEtAl-2013-CSC,
  Heinz-Idsardi-2013-WCDRADL, rogers-lambert-2019-classes,
  Lambert2023}.

Second, MLRegTest is much more comprehensive and makes more
fine-grained distinctions than previous work. For example, consider
\citet{Tomita1982}. There were seven Tomita languages altogether, the
alphabet size was restricted to two symbols, and the largest DFA has
four states. MLRegTest improves each of these metrics and so it is
much more comprehensive. There are 1,800 languages; 3 alphabet sizes
are used (4, 16, and 64); and the minimum, maximum, median, mean and
standard deviations of the sizes of the minimal DFA and their
syntactic monoids are shown in Table~\ref{tab:sizes}.
\begin{table}[ht]
  \centering
  \begin{tabular}{llllll}
    \toprule
    Type of Machine       & min & max  & median & mean  & s.d.   \\
    \midrule
    Minimal DFA           & 2   & 613  & 11     & 23.45 & 53.24  \\
    Monoid of Minimal DFA & 2   & 4229 & 51     & 155.4 & 329.89 \\
    \bottomrule
  \end{tabular}
  \caption{Summary statistics of the numbers of states in the minimal
    deterministic acceptors and their syntactic monoids of the
    languages in MLRegTest.}
  \label{tab:sizes}
\end{table}

Another example comes from recent work which studied transformer and
LSTM performance on regular languages organized by their dot-depth
\citep{bhattamishra-etal-2020-ability}. They consider 30 regular
languages whose complexity varies according to where they fall on the
dot-depth hierarchy. Like the classes presented here, the dot-depth
classes are mathematically well-understood. However, the simplest
class they consider, the dot-depth one class, is defined nearly
identically to what we call the Piecewise Local Testable (PLT) class
\citep{Lambert2022}, and MLRegTest considers hundreds of languages
from 11 subclasses of PLT. In other words, there are many
mathematically natural (as evidenced by their many characterizations)
classes of languages within the simplest class considered by
Bhattamishra et al.\@ which are not distinguished in their study, but
which MLRegTest does distinguish. Furthermore, these classes are also
motivated by linguistic and cognitive considerations
\citep{Rogers-HeinzEtAl-2010-LPTSS, Heinz-RawalEtAl-2011-TSLCP,
  Rogers-HeinzEtAl-2013-CSC, Heinz-2018-CNPG}. To our knowledge,
MLRegTest is the most comprehensive, fine-grained suite of artificial
regular languages ever constructed.

\section{Languages} 
\label{sec:lgs}
This section describes the 16 classes of formal languages from which
the 1,800 languages were drawn. The first part discusses the classes
themselves, and the second part discusses how we designed the 1,800
languages in the dataset.

\subsection{Subregular Formal Languages}

\begin{figure}
  \centering
  \hspace{-7mm}\begin{tikzpicture}[scale=0.67]
  \footnotesize
  \tikzset{
    co/.style={  
      rectangle split,
      rectangle split parts=2,
      rectangle split horizontal,
      text centered
    },
    pl/.style={  
      text centered
    },
  }

  \draw[rounded corners,thick,lightgray,
  fill=yellow,fill opacity=.2] (-5.75,11) rectangle (-3.25,0.5);
  \draw[rounded corners,thick,lightgray,
  fill=blue,fill opacity=.2] (-3.25,11) rectangle (-0.75,0.5);
  \draw[rounded corners,thick,lightgray,
  fill=green,fill opacity=.2] (-0.75,11) rectangle (2,0.5);
  \draw[rounded corners,thick,lightgray,
  fill=orange,fill opacity=.2] (2,11) rectangle (4.5,0.5);
  \draw[rounded corners,thick,lightgray,
  fill=red,fill opacity=.2] (4.5,11) rectangle (7,0.5);

  \node[] (sl)   at (-5.25,3)  {\textbf{SL}};
  \node[] (cosl) at (-4.25,2.25)  {\textit{coSL}};

  \node[] (tsl)   at (-2.5,3)  {\textbf{TSL}};
  \node[] (tcosl) at (-1.5,2.25)  {\textit{TcoSL}};

  \node[pl] (lt)   at (-4.5,5.25) {LT};
  \node[pl] (ltt)  at (-4.5,7.75)  {LTT};

  \node[pl] (tlt)  at (-2,5.25)  {TLT};
  \node[pl] (tltt) at (-2,7.75)   {TLTT};

  \node[pl] (lp)   at (0.625,5.25)   {PLT};
  \node[pl] (tlp)  at (3.25,5.25)   {TPLT};

  \node[] (sp)   at (5.25,3)  {\textbf{SP}};
  \node[] (cosp)   at (6.25,2.25)  {\textit{coSP}};

  \node[pl] (pt)   at (5.75,5.25) {PT};
  \node[pl] (sf)   at (3.25,7.75)   {~S~t~a~r~~~F~r~e~e};

  \draw[thick,rounded corners] (-0.7,6.6) rectangle (6.95,8.915);

  \draw[thick,rounded corners] (-5.75,9.1) rectangle (6.95,10.95);

  \node[] (reg) at (0.625,10.5) {R~e~g~u~l~a~r};
  \node[] (zp)   at (0.625,9.4) {$\mathbb{Z}_p$};

  \path[->]
  (reg)  edge[] (zp)
  (reg.east)  edge[bend left] (sf.north)

  (sf.south)   edge[] (tlp)
  (tlp)  edge[] (lp)
  (lp) edge[out=160,in=315] (ltt)
  (lp) edge[bend right] (pt)
  (tlp) edge[out=160,in=300] (tltt)

  (tltt) edge[] (ltt)
  (tltt) edge[] (tlt)
  (tlt)  edge[] (lt)

  (tlt.south)  edge[] (tsl)
  (tlt.south)  edge[] (tcosl)
  (ltt)  edge[] (lt)
  (lt.south)   edge[] (sl)
  (lt.south)   edge[] (cosl)
  (pt.south)   edge[] (sp)
  (pt.south)   edge[] (cosp)

  (tsl) edge[] (sl)
  (tcosl) edge[] (cosl)
  ;



  \draw[blue,thick, dashed] (-9.5,4) -- (7.5,4);
  \draw[blue,thick, dashed] (-9.5,6.5) -- (7.5,6.5);
  \draw[blue,thick, dashed] (-9.5,9) -- (7.5,9);

  \node[text=blue,draw=none,align=center] at (-8,10)    (mso) {Monadic Second\\ Order Logic};
  \node[text=blue,draw=none,align=center] at (-8,7.75) (fo)  {First Order\\ Logic};
  \node[text=blue,draw=none,align=center] at (-8,5.25)  (pl)  {Propositional\\ Logic};
  \node[text=blue,draw=none,align=center] at (-8,3)   (cnl) {Conjunctions of\\ Negative Literals\\ (boldface)};
  \node[text=blue,draw=none,align=center] at (-8,1)     (dpl) {Disjunctions of\\ Positive Literals\\ (italics)};

  \node[text=red,draw=none] at (-4.5,0.95)  (succ) {successor};
  \node[text=red,draw=none,align=center] at (-2,1.25)  (tsucc) {tier\\[-0.5ex] successor};
  \node[text=red,draw=none,align=center] at (0.6,1.15)  (scpr) {successor,\\[-0.5ex] precedence};
  \node[text=red,draw=none,align=center] at (3.25,1.45)  (tscpr) {tier\\[-0.5ex] successor,\\[-0.5ex] precedence};
  \node[text=red,draw=none] at (5.75,1) (prec) {precedence};


\end{tikzpicture}
  \caption{Regular and subregular classes of formal languages
    organized by logical language and ordering relation(s).
    An arrow from class \(A\) to class \(B\)
    indicates that class \(A\) is a proper superclass
    of class \(B\).}
  \label{fig:classes}
\end{figure}
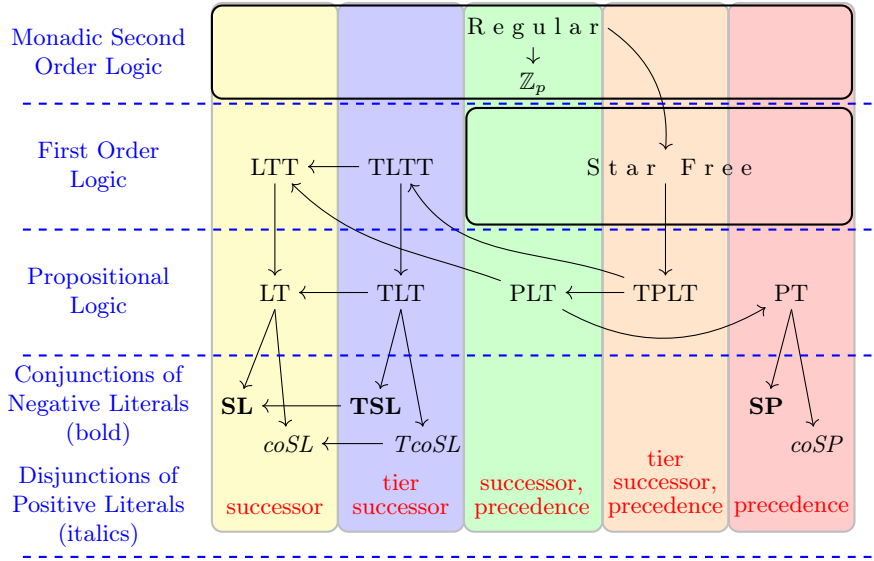

An underlying theme to the 16 classes we consider is the notion of
string containment. This notion can ultimately be dissected along two
dimensions, logical power and representation, using the tools of
mathematical logic and model theory \citep{Enderton2001,Libkin2004}.
Figure~\ref{fig:classes} shows the 16 classes considered in this paper
with arrows indicating proper subset relationships among them.
The vertical axis in Figure~\ref{fig:classes} is organized in terms of
different logics, with horizontal blue-dashed lines indicating leaps
in logical power. The horizontal axis in Figure~\ref{fig:classes} is
organized according to the primitive represenational elements in the
logical languages. In model-theoretic parlance, these representational
choices constitute what is called the \emph{model signature}. The
horizontal axis is not ordered in terms of increasing power like the
vertical axis.

It is worth mentioning that for each class $C$ in
Figure~\ref{fig:classes}, there is an algorithm which can take any
finite-state acceptor and decides whether the language recognized by
that acceptor belongs to $C$ or not. Many of these algorithms are
based on the algebraic properties of these classes
\citep{Pin2021}. These algorithms have been implemented in the
software packages The Language Toolkit and Amalgam \citep{Lambert2022,
  Lambert2024}.\footnote{Available at
  \url{https://hackage.haskell.org/package/language-toolkit} and
  \url{https://github.com/vvulpes0/amalgam}~.}

We explain these classes of languages by first considering languages
defined via the containment of substrings (the ``successor'' column),
and then by exploring different logics based on this notion.  We then
expand the notion of substring containment to subsequences (the
``precedence'' column) and then to substrings on projected tiers
(``tier-successor'') and then to their combinations.

\subsubsection{The Local Family}

\begin{figure}[ht]
  \centering
  \begin{tikzpicture}
  [->,shorten >=1pt,node distance=20mm,semithick]
  \tikzstyle{every state}=[fill=white,draw=black,text=black,inner sep=3pt,minimum size=0pt]

  \node[initial,initial text={},state] (q0) {0};
  \node[state] (q1) [right of = q0] {1};
  \node[state,accepting,double distance=1pt] (q2) [right of = q1] {2};

  \path[->,auto,font=\footnotesize]
  (q0) edge[out=135,in=45,min distance=10mm,looseness=5] node {b,c,d,e} (q0)
  (q0) edge[bend left] node {a} (q1)
  (q1) edge[] node {a} (q2)
  (q1) edge[bend left] node {b,c,d,e} (q0)
  (q2) edge[out=135,in=45,min distance=10mm,looseness=5] node {a,b,c,d,e} (q2)
  ;
\end{tikzpicture}
  \caption{A finite-state acceptor recognizing the language of all and
    only those strings which contain the $aa$ substring.}
  \label{fig:sl-aa}
\end{figure}
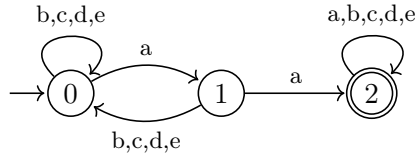

For $w\in\Sigma^*$, the regular expression $\Sigma^*w\Sigma^*$
represents the set of all and only those strings which contain
$w$ as a substring. Let $C(w)=\Sigma^*w\Sigma^*$. As an example,
Figure~\ref{fig:sl-aa} shows a finite-state acceptor which recognizes
$C(aa)$.
One class of languages we consider can be defined by taking finitely
many strings $w_1,w_2, \ldots w_n$ and constructing the union of the
languages which contain them.
\begin{equation}
\bigcup_{\mathclap{1\leq i \leq n}} C(w_i)\label{eq:1}
\end{equation}
To decide whether a string $x$ belongs to such a language requires
identifying whether $x$ contains any one of the substrings
$w_1,w_2, \ldots w_n$. If it does then $x$ belongs to the language, and
otherwise it does not. The $w_i$ are ``licensing'' substrings, and
strings must possess at least one licensing substring.

The complements of the aforementioned languages present another class
of languages.  In this case, by DeMorgan's law, any complement
language would have the form shown in Equation~\ref{eq:2}, where
$\overline{C(w_i)}$ indicates the complement of $C(w_i)$; that is, the
set of all strings which do not contain $w_i$ as a substring.
\begin{equation}
\bigcap_{\mathclap{1\leq i \leq n}} \overline{C(w_i)}\label{eq:2}
\end{equation}
To decide whether a string $x$ belongs to such a
language also requires identifying whether $x$ contains any one of the
substrings $w_1,w_2, \ldots w_n$. If it does then $x$ does not belong to
the language, and otherwise it does. Here, the $w_i$ are ``forbidden''
substrings, and strings must not possess any forbidden substring.

Historically, the latter class of languages was studied first, and it
is called the Strictly Local (SL) class, or sometimes Locally
Testable in the Strict Sense \citep{McNaughtonPapert1971}. Following
\citet{rogers-lambert-2019-classes}, we call the %
former class Complements of Strictly Local (coSL). It can be argued
that neither of these classes makes use of long-term
dependencies. This is because there are only finitely many $w_i$ and
so there is a longest one of length $k$.  Therefore, deciding whether
a string $x$ contains any $w_i$ comes down to scanning $x$ with
windows of size $k$. All the information needed to decide string
membership is local within bounded windows of size $k$.

From a logical perspective, the SL class can be understood as the
conjunctions of negative literals and the coSL class can be understood
as the disjunctions of positive literals. Here positive literals are
strings $w$ and they are interpreted as $C(w)$. A negative literal is
$\neg w$, which is interpreted as $\overline{C(w_i)}$. In this way,
we obtain a direct translation of these language classes into
particular Boolean expressions
\citep{rogers-lambert-2019-classes}. Specifically, in terms of these
logical expressions SL languages will have the logical form shown in
Equation~\ref{eq:3}, and coSL languages will have the logical form
shown in Equation~\ref{eq:4}.
\begin{equation}
 \bigwedge_{\mathclap{1\leq i \leq n}} \neg w_i\label{eq:3}
\end{equation}
\begin{equation}
 \bigvee_{\mathclap{1\leq i \leq n}} w_i \label{eq:4}
\end{equation}

Long-distance dependencies appear when the logical formalism
introduced above is generalized to any Boolean expression over strings
as literals. For example, the Boolean expression in Equation~\ref{eq:5}
would be interpreted as the set of strings $x$ such that if $x$
contains the substring $aa$ then it also contains the substring
$ab$.
\begin{equation}
aa \rightarrow ab\label{eq:5}
\end{equation}
Note in this language, the substrings $aa$ and $ab$ do not need
to be adjacent, or even in any particular order. For example, strings
$aab$ and $c^{10}abc^{20}aac^{20}$ belong to this language and strings
$baa$ and $c^{10}aac^{20}ac^{20}$ do not. This class of languages is
called the Locally Testable (LT) class
\citep{McNaughtonPapert1971}. Deciding whether a string $x$ belongs to
a LT language requires keeping track of the substrings that occur in
$k$-sized windows in $x$ \citep{Rogers-HeinzEtAl-2013-CSC}. Again
because the Boolean expression is of finite length, there is a longest
literal $w_i$ of length $k$.

The Locally Threshold Testable (LTT) class of languages generalizes
the LT class. Deciding whether a string $x$ belongs to a LTT language
requires keeping track of how many substrings there are, counting them
up to some threshold $t$, that occur in $k$-sized windows in $x$
\citep{RogersPullum2011, Rogers-HeinzEtAl-2013-CSC}. LTT is a
superclass of LT. In fact, LT is the subclass of LTT where the
threshold $t$ equals 1.

An example of a LTT language is one that requires there to be at least
two $aa$ substrings in a word. In this language, for all $n$,
$b^naab^naab^n$ belongs to this language but $b^naab^n$ does not. One
can prove that there is no Boolean expression which represents this
language and so it is not LT. But it is LTT, where the threshold
equals 2, and so it is possible to distinguish between 0, 1 and 2 or
more occurrences of substrings of length $k$. Note this is a kind of
long-term dependency distinct in kind from the ones presented in the
LT class.

First Order (FO) logic is a more powerful logic than Boolean logic. FO
logic includes universal and existential quantification over elements
in a structure. Defining a FO logical language, then, requires clarity
about the structures being described. Model theory provides a way to
talk about mathematical structures and the relations that make up such
structures \citep{Enderton2001,Hedman2004}. Strings are one such
mathematical structure and are well studied in this way. In model-theoretic
representations of strings, the successor relation is one way in which
the order of the elements can be encoded, and its usage yields the
notion of substring that was used in defining LT, SL, and coSL. (Other
possible relations for strings are discussed later.)
\citet{Thomas1982} showed that the class of formal languages definable
in First Order logic with the successor relation is exactly the
Locally Threshold Testable (LTT) class.

The last move in logical power is to move from FO logic to Monadic
Second Order (MSO) logic. MSO logic extends FO logic by additionally
allowing quantification over sets of elements in structures.
\cite{Buchi1960} established that the languages definable with
finite-state acceptors are exactly the ones definable in MSO logic
with the successor relation. Readers are referred to
\citet{Thomas1997} for more details. These are thus a proper superset
of LTT.

Parity languages are examples of regular languages that are not LTT. A
parity language is a language which counts modulo $n$. For example, a
language that requires there to be an even number of $a$s in strings
is an example of a parity language and is shown in
Figure~\ref{fig:even-a}.
\begin{figure}[ht]
  \centering
  \begin{tikzpicture}
  [->,shorten >=1pt,node distance=20mm,semithick]
  \tikzstyle{every state}=[fill=white,draw=black,text=black,inner sep=3pt,minimum size=0pt]

  \node[initial,initial text={},state,accepting,double distance=1pt] (q0) {0};
  \node[state] (q1) [right of = q0] {1};

  \path[->,auto,font=\footnotesize]
  (q0) edge[out=135,in=45,min distance=10mm,looseness=5] node {b,c,d,e} (q0)
  (q0) edge[bend left] node {a} (q1)
  (q1) edge[bend left] node {a} (q0)
  (q1) edge[out=135,in=45,min distance=10mm,looseness=5] node {b,c,d,e} (q1)
  ;
\end{tikzpicture}
  \caption{A finite-state acceptor recognizing the language of all and
    only those strings which contain an even number of $a$s.}
  \label{fig:even-a}
\end{figure}
Pure modulo-counting with a prime
modulus forms a class we call $\mathbb{Z}_p$, named for the algebraic
groups ($\mathbb{Z}/p\mathbb{Z}$) that their automata invoke.

These classes, SL, coSL, LT, LTT and Regular are shown in the leftmost
column labeled ``successor'' in Figure~\ref{fig:classes}. The class
$\mathbb{Z}_p$ is a proper subset of the Regular languages and
disjoint from these others.

\subsubsection{The Piecewise Family}

We next modify the notion of containment from substring to subsequence
(the rightmost column in Figure~\ref{fig:classes}). For
$w=a_1a_2\ldots a_n\in\Sigma^*$, the regular expression
$\Sigma^*a_1\Sigma^*a_2\Sigma^*\ldots \Sigma^*a_n\Sigma^*$ represents
the set of all and only those strings which contain $w$ as a
subsequence. Let
$C_<(w)=\Sigma^*a_1\Sigma^*a_2\Sigma^*\ldots \Sigma^*a_n\Sigma^*$. The
choice of subscript for $C_<$ is motivated by the fact that the
precedence relation ($<$) is used to represent the order of elements
in a string in place of the successor relation in model-theoretic
treatments \citep{McNaughtonPapert1971,Rogers-HeinzEtAl-2013-CSC}.

\begin{figure}[ht]
  \centering
  \begin{tikzpicture}
  [->,shorten >=1pt,node distance=20mm,semithick]
  \tikzstyle{every state}=[fill=white,draw=black,text=black,inner sep=3pt,minimum size=0pt]

  \node[initial,initial text={},state] (q0) {0};
  \node[state] (q1) [right of = q0] {1};
  \node[state,accepting,double distance=1pt] (q2) [right of = q1] {2};

  \path[->,auto,font=\footnotesize]
  (q0) edge[out=135,in=45,min distance=10mm,looseness=5] node {b,c,d,e} (q0)
  (q0) edge[] node {a} (q1)
  (q1) edge[] node {a} (q2)
  (q1) edge[out=135,in=45,min distance=10mm,looseness=5] node {b,c,d,e} (q1)
  (q2) edge[out=135,in=45,min distance=10mm,looseness=5] node {a,b,c,d,e} (q2)
  ;
\end{tikzpicture}
  \caption{A finite-state acceptor recognizing the language of all and
    only those strings which contain the $aa$ subsequence.}
  \label{fig:sp-aa}
\end{figure}
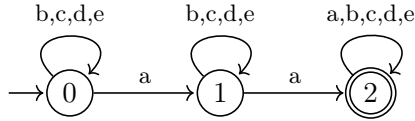

As an example, Figure~\ref{fig:sp-aa} shows a finite-state acceptor
which recognizes $C_<(aa)$.
Words like $c^{20}ac^{20}ac^{20}$ belong to this language but words
like $c^{20}ac^{20}bc^{20}$ do not.

The ``Piecewise'' families of languages can then be constructed
exactly as before. The Strictly Piecewise (SP) class of languages is
defined by taking finitely many strings\linebreak $w_1, w_2, \ldots w_n$ and
constructing the intersection of the languages which do not contain
these strings as subsequences (Equation~\ref{eq:6}).
\begin{equation}
\bigcap_{\mathclap{1\leq i \leq n}} \overline{C_<(w_i)}\label{eq:6}
\end{equation}
To decide whether a string $x$ belongs to such a language requires
identifying whether $x$ contains any one of the subsequences
$w_1,w_2, \ldots w_n$. If it does not contain one then $x$ belongs to
the language, and if does contain one then it does not. The $w_i$ are
``forbidden'' subsequences, and strings must not possess any forbidden
subsequences \citep{Rogers-HeinzEtAl-2010-LPTSS}. Using the Boolean
expressions mentioned previously, each SP language can be expressed as
shown in Equation~\ref{eq:7} with where each $w_i$ is interpreted as
the language containing $w_i$ as a subsequence
\citep{Rogers-HeinzEtAl-2013-CSC}.
\begin{equation}
 \bigwedge_{\mathclap{1\leq i \leq n}} \neg w_i \label{eq:7}
\end{equation}

Similarly, the Complement of Strictly Piecewise (coSP) class of
languages is defined by taking finitely many strings
$w_1,w_2, \ldots w_n$ and constructing the union of the languages
which contain these strings as subsequences.
\begin{equation}
\bigcup_{\mathclap{1\leq i \leq n}} C_<(w_i)\label{eq:8}
\end{equation}
Here, the $w_i$ are ``licensing'' subsequences, and to belong to the
language, a string must possess at least one licensing subsequence
\citep{rogers-lambert-2019-classes}.  It follows that the coSP
languages can be expressed with the Boolean expression shown in
Equation~\ref{eq:9}
\begin{equation}
  \bigvee_{\mathclap{1\leq i \leq n}} w_i\label{eq:9}
\end{equation}
The SP and coSP language classes
are incomparable with the LTT class. In other words, they generally
encode different kinds of long-term dependencies than those in the
LTT and LT languages.

Like the LT class, the Piecewise Testable (PT) class of languages
\citep{Simon1975} is characterized with any Boolean expression over
literals. The literals are now interpreted as containment by
subsequence. For example, the Boolean expression in Equation~\ref{eq:10}
would be interpreted as the set of strings $x$ such that if $x$
contains the subsequence $aa$ then it also contains the subsequence
$ab$.
\begin{equation}
aa \rightarrow ab\label{eq:10}
\end{equation}
For example, strings $aba$ and $c^{10}ac^{20}ac^{20}bc^{20}$ belong to
this language and strings $baa$ and $c^{10}ac^{20}ac^{20}$ do not. It
follows that deciding whether a string $x$ belongs to a PT language
whose longest literal is of length $k$ requires keeping track of the
subsequences of size $k$ that occur in $x$
\citep{Rogers-HeinzEtAl-2013-CSC}.

The logical language obtained by combining the precedence relation
with FO logic yields formulas whose corresponding languages form
exactly the Star-Free (SF) class of languages
\citep{McNaughtonPapert1971}. The name Star-Free comes from one of the
first definitions of this class in terms of star-free regular
expressions; that is languages describable with the base cases
$a\in\Sigma$, $\emptyset$, and $\epsilon$ (the empty string), and
operations union, intersection, concatenation, and complement with
respect to $\Sigma^*$, but crucially the Kleene star operation is
omitted. This celebrated result, along with other characterizations,
is due to McNaughton and Papert \citep{McNaughtonPapert1971}.

That the SF class properly contains the LTT class follows from the
fact that the successor relation is FO definable with precedence, but
not vice versa. To define successor with precedence, consider the
following: $s(x,y) := x < y \wedge \neg \exists z [ x < z <
y]$. \citet{Thomas1997} provides a proof that precedence is not
definable with successor. A concrete example of a SF language that is
neither LTT nor PT is the language obtained by concatenating all words
which end with the symbol $a$ with all words that do not contain a
$bc$ substring. Formally, with an alphabet $\Sigma=\{a, b, c, d\}$,
this language can be expressed as $\Sigma^* a\overline{C(bc)}$.

When the precedence relation is combined with MSO logic, exactly the
class of regular languages is obtained again. This is because precedence
is MSO definable with successor and vice versa. The classes, SP,
coSP, Star-Free, and Regular are shown in the rightmost column labeled
``precedence'' in Figure~\ref{fig:classes}.

\subsubsection{The Tier-Local Family}

The tier-local family of classes introduces yet another kind of
long-distance dependency that similarly interacts with the logical
languages already introduced. In this family of language classes,
the notion of containment
involves a sequence of `salient' symbols
which is contained when it appears as a substring
when non-salient symbols are ignored.
The set of salient symbols is called the ``tier'' and is some subset
$T\subseteq\Sigma$ \citep{Heinz-RawalEtAl-2011-TSLCP,
  Lambert2023}. For example, if $\Sigma=\{a,b,c,d,e\}$ and $T=\{a,e\}$
and $w=daceba$ then the string on tier $T$ is $aea$. The notion of
``contains the substring on the tier'' can be generally expressed as
follows. For $w=a_1a_2\ldots a_n\in T^*$, the regular expression
$\Sigma^*a_1\overline{T}^*a_2\overline{T}^*\ldots
\overline{T}^*a_n\Sigma^*$ where $\overline{T}^*=(\Sigma-T)^*$
represents the set of all and only those strings which contain $w$ as
a substring when all the non-tier symbols are removed.

\begin{figure}[ht]
  \centering
  \begin{tikzpicture}
  [->,shorten >=1pt,node distance=20mm,semithick]
  \tikzstyle{every state}=[fill=white,draw=black,text=black,inner sep=3pt,minimum size=0pt]

  \node[initial,initial text={},state] (q0) {0};
  \node[state] (q1) [right of = q0] {1};
  \node[state,accepting,double distance=1pt] (q2) [right of = q1] {2};

  \path[->,auto,font=\footnotesize]
  (q0) edge[out=135,in=45,min distance=10mm,looseness=5] node {b,c,d,e} (q0)
  (q0) edge[bend left] node {a} (q1)
  (q1) edge[] node {a} (q2)
  (q1) edge[bend left] node {e} (q0)
  (q1) edge[out=135,in=45,min distance=10mm,looseness=5] node {b,c,d} (q1)
  (q2) edge[out=135,in=45,min distance=10mm,looseness=5] node {a,b,c,d,e} (q2)
  ;
\end{tikzpicture}
  \caption{A finite-state acceptor recognizing the language of all and
    only those strings which contain $aa$ as a substring on the
    $\{a,e\}$ tier.}
  \label{fig:tsl-aa}
\end{figure}
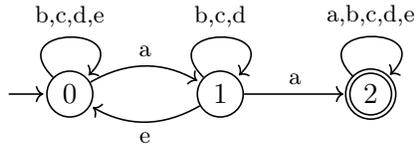

For example, Figure~\ref{fig:tsl-aa} shows a finite-state acceptor
which recognizes $C_T(aa)$ where $T=\{a,e\}$.
This is the language which must contain the substring $aa$ on the
$\{a,e\}$ tier. So words like $c^{20}ac^{20}ac^{20}ec^{20}$ belong to
this language but words like $c^{20}ac^{20}ec^{20}ac^{20}$ do not. The
former has the string $aae$ on this tier whereas the latter has $aea$.

Letting
$C_T(w)=\Sigma^*a_1\overline{T}^*a_2\overline{T}^*\ldots
\overline{T}^*a_n\Sigma^*$, we can define the Tier Strictly Local
(TSL) and Complement of Tier Strictly Local (TcoSL) classes using
conjunctive and disjunctive fragments of Boolean logic analogously to
the SL, SP, coSL, and coSP classes. Similarly the use of propositional
logic will characterize the Tier Locally Testable (TLT) class. Note
for all languages within all of these classes, the tier $T$ remains
invariant. For example the formula shown in Equation~\ref{eq:11} will
be interpreted as $\overline{C_T(ae)} \cap \overline{C_T(ae)}$.
\begin{equation}
 \neg ae \wedge \neg ea\label{eq:11}
\end{equation}
So each term in the formula is interpreted with respect to the same
tier $T$. The tier $T$ does vary across these formal languages within
the class, but not within individual languages. (See
\citet{AksenovaDeshmukh2018} and \citet{Lambert2022} for research on
languages incorporating multiple tiers.)

When we move to FO logic, instead of the successor or precedence
relations, the order relation is the tier-successor relation (specific
again to some $T$). This representation of strings combined with FO
logic yields the TLTT class. If MSO logic is used with the tier-successor
relation, the class of regular languages is again obtained.

When $T=\Sigma$, every symbol is salient and this
special case reduces to Local family of languages. It follows that TSL
is a proper superset of SL, TcoSL is a proper superset of coSL, and so
on as shown in Figure~\ref{fig:classes}.

Interestingly, projecting salient symbols and then checking for
subsequence containment (precedence) does not lead to more expressive
classes. In other words, everything tier-precedence can do, precedence
can already do.

\subsubsection{More Than One Order Relation}

Including the precedence relation with either the successor relation
or the tier-successor relation in the model-theoretic representation
yields more expressive power than any of these relations on its own
with propositional logics. For instance when both the precedence and
successor relations are included, the literals in the logical language
refer to sequences which are contiguous at some points and
discontiguous at others. For example, let $\scr$ denote the successor
relation and $<$ the precedence relation. The substring $aaa$ would
now be written $a\scr a\scr a$ and the subsequence $aaa$ would now be
written $a< a< a$. A literal such as $a\scr a< a \scr b$ now denotes
the set of all strings which contain a substring $aa$ which precedes a
substring $ab$.

In this way, combining adjacency and general precedence at the
propositional level allows local and long-distance conditions to
co-occur within a single constraint. This is the Piecewise Locally
Testable (PLT) class. Similarly, The Tier Piecewise Locally Testable
and (TPLT) combines tier-adjacency and general precedence for a
similar purpose. TPLT properly includes PLT. Interestingly LTT
$\subsetneq$ PLT and TLTT $\subsetneq$ TPLT
\citep{rogers-lambert-2019-classes,Lambert2022}.

If FO logic is used, then the addition of successor or tier-successor
to precedence does not increase the expressive power of the logical
language, which yields the Star-Free languages.

The ``strict'' counterparts of PLT and TPLT also exist. They are
omitted from this study because we are not aware of any implementation
deciding membership in them, contra the situation for PLT and TPLT
\citep{Lambert2022}.

\subsection{Summary}

The classes presented here identify several types of formal languages.
Among the simplest are the SL languages which forbid specific
substrings from occurring.  This kind of constraint, based on a
conjunction of negative literals, specifies a local dependency.  Its
complement (coSL), a disjunction of positive literals, would be a
different sort of local dependency, where substrings license, rather
than forbid, strings in the language.

The other classes enable different sorts of long-term
dependencies. For example, the Piecewise classes encode long-distance
dependencies based on subsequences, and the Tier-Local classes encode
long-distance dependencies based on strings of salient
symbols. Consequently, the SP languages forbid subsequences from
occurring, and the Tier Strictly Local languages forbid substrings from
occurring on tiers of salient symbols.

We call these different kinds of strings---substrings, subsequences,
projected substrings on tiers, and combinations
thereof---\emph{factors}.  Adding arbitrary Boolean combinations
results in a full propositional logic, which allows conditional
constraints so that the presence or absence of a particular set of
factors can trigger the enforcement of another local dependency. These
are the Testable languages (LT, PT, TLT, PLT, and TPLT).  FO logic
lets one count instances of factors up to some threshold (LTT, TLTT)
and MSO logic lets one count them relative to some modulus (Regular).

Finally it is worth mentioning that once a model signature has been
fixed, any class at or below the propositional level has an associated
parameterized learner that converges with complete accuracy without
any negative data at all and whose sample complexity is relatively
small \citep{Lambert+2021-TESRL}.

\subsection{The Languages in MLRegTest}\label{subsec:mlregtest}

MLRegTest contains representations of 1,800 languages drawn from the 16
classes described above. This section explains how those 1,800
languages were constructed, and the design choices that went into
their construction.

An important design goal was to ensure that each language in MLRegTest
counts as a representative of a single class. Since classes may fully
or partially include other classes, a typical formal language actually
belongs to more than one class. For example, every SL language is LT
but not vice versa. Consequently, we designed MLRegTest such that a
language $L$ counts as a representative of a class $X$ provided that
$L$ belongs to class $X$, and $L$ does not belong to any class $Y$ which
is a subset of, or incomparable with, $X$. Following this principle, a
SL language could count as a representative of the SL class, but not
the LT class. Furthermore, the languages representative of the LT
class will not belong to SL, coSL, TSL, TcoSL, or PT. Henceforth, when
referring to languages and classes in MLRegTest and we write ``$L$
belongs to class $X$'', or ``the languages in class $X$'', ``language
$L$ from class $X$'', or anything similar, we mean the language $L$
counts as representative of the class $X$ in the manner described
here.

One caveat with the above approach is that it presupposes that the
inherent complexity of a class will be demonstrated with the languages
which are representative of the class in the above sense. While we
believe this is a reasonable position to adopt, the experiments
presented later provide some evidence that it is not entirely the case
(see \S\ref{sec:res:lgclass}).

After presenting some additional parameters of our design, we explain
how we algorithmically verified that we achieved the aforementioned
design goal. However, the specific choices of parameter values was
influenced by our ability to conduct verification. In particular, we
often adopted numbers that made verification possible. Next we discuss
those parameterizations.

For each class, we developed ten base patterns. For example, in the SL
class, one base pattern only forbids strings containing $a^k$, for the
symbol $a$ in the alphabet and for some $k$. Another pattern forbids
$(ab)^{k/2}$ when $k$ is even. These base patterns are then actualized
by specifying both the alphabet and the value $k$, dimensions of
variation to which we now turn. The languages in the Reg class were
obtained by intersecting a language in $\mathbb{Z}_p$ with a language
in a class other than $\mathbb{Z}_p$ or Reg.

For all language classes, the base patterns were embedded in three
alphabet sizes \{4, 16, 64\}. The alphabets are nested. The sizes were
chosen to grow exponentially. Specifically, the alphabets were the
first 4, 16, or 64 letters of the sequence
$\langle$\texttt{abcdefghijklmnopqrstuvwxyz\linebreak
ABCDEFGHIJKLMNOPQRSTUVWXYZ\'a\`a\v{a}\'e\`e\v{e}\'o\`o\v{o}\'u\`u\v{u}}%
$\rangle$.

One of the key properties of the languages in all the classes, except
for the SF, $\mathbb{Z}_p$, and Reg classes, is the window size $k$,
which corresponds to the length of the longest literal (string) in the
logical expression describing the pattern. We considered three $k$
values: \{2,4,6\}.

The language classes SL, coSL, SP, coSP, LT, PLT, and PT only vary
across the dimensions of base, alphabet, and $k$ value. Therefore, we
constructed 90 languages in each of these classes (10 bases $\times$ 3
alphabets $\times$ 3 $k$ values).

The SF, $\mathbb{Z}_p$, and Reg classes are not specified in terms of $k$ value,
tiers, or thresholds. Therefore, there are only 30 languages in these
classes (10 bases $\times$ 3 alphabets). 

An additional parameter of variation for the classes TSL, TcoSL, TLT,
and TPLT is the number of salient symbols (those that project onto the
tier). Because having more or fewer symbols be salient might affect
learning difficulty, we provided two tier sizes for alphabets 16 and
64. For the alphabet of size 16, the tier sizes were \{4, 7\}; and for
the alphabet of size 64, the tier sizes were \{6, 11\}.  When the
alphabet was of size 4, we only included one tier of size \{3\}
because we could not otherwise easily construct languages that we
could verify as representatives of TLT and TPLT.

The LTT class does not have a tier, but it does have an additional
parameter, which is the counting threshold. We considered three
thresholds \{2,3,5\} but they are not equally represented in
MLRegTest. Instead, they occur in a 3:2:1 ratio so that we have 90
languages with threshold 2, 60 with threshold 3, and 30 with threshold
5. Consequently, there were a total of 180 languages in LTT.

Finally the TLTT class has both a tier and a threshold. The thresholds
were chosen the same way as the LTT class. Also, the tiers were chosen
the same way as the other tier classes. Therefore, there was a total
of 300 TLTT languages (60 languages for each alphabet size/tier size
combination, of which there are 5: 4/3, 16/7, 16/4, 64/11, 64/6). Within
each group of 60, there are 30 languages with threshold 2, 20 with
threshold 3, and 10 with threshold 5.

Tables~\ref{tab:lgs1} and \ref{tab:lgs2} summarize the design
parameters that led to the construction of 1,800 languages in
MLRegTest, as well as show the number of languages in each class.
\begin{table}[h]
  \centering
  \begin{tabular}{lllllll}
    \toprule
    class          & bases & alphabets & windows & thresholds & total                     \\
    \midrule
    SL             & 10    & 3         & 3       &            & 90                        \\
    coSL           & 10    & 3         & 3       &            & 90                        \\
    SP             & 10    & 3         & 3       &            & 90                        \\
    coSP           & 10    & 3         & 3       &            & 90                        \\
    \midrule
    LT             & 10    & 3         & 3       &            & 90                        \\
    PLT            & 10    & 3         & 3       &            & 90                        \\
    PT             & 10    & 3         & 3       &            & 90                        \\
    \midrule
    LTT            & 10    & 3         & 3       & 2(3)*      & 180                       \\
    SF             & 10    & 3         &         &            & 30                        \\
    $\mathbb{Z}_p$ & 10    & 3         &         &            & 30                        \\
    Reg            & 10    & 3         &         &            & 30                        \\
    \midrule
    total          &       &           &         &            & \textbf{900}              \\
    \bottomrule
  \end{tabular}
  \caption{A summary of the number of languages without tiers in each
    class and the dimensions along which they vary. The asterisk
    indicates that while there were actually 3 thresholds, since they
    occur in 3:2:1 ratio, they only doubled the number of languages.}
  \label{tab:lgs1}
\end{table}

\begin{table}[h]
  \centering
  \begin{tabular}{lllllll}
    \toprule
    class          & bases & alphabets & windows & tiers      & thresholds & total        \\
    \midrule
    TSL            & 10    & 1         & 3       & 1          &            & 30           \\
                   & 10    & 2         & 3       & 2          &            & 120          \\
    \midrule
    TcoSL          & 10    & 1         & 3       & 1          &            & 30           \\
                   & 10    & 2         & 3       & 2          &            & 120          \\
    \midrule
    TLT            & 10    & 1         & 3       & 1          &            & 30           \\
                   & 10    & 2         & 3       & 2          &            & 120          \\
    \midrule
    TPLT           & 10    & 1         & 3       & 1          &            & 30           \\
                   & 10    & 2         & 3       & 2          &            & 120          \\
    \midrule
    TLTT           & 10    & 1         & 3       & 1          & 2(3)*      & 60           \\
                   & 10    & 2         & 3       & 2          & 2(3)*      & 240          \\
    \midrule
    total          &       &           &         &            &            & \textbf{900} \\
    \bottomrule
  \end{tabular}
  \caption{A summary of the number of languages with tiers in each
    class and the dimensions along which they vary. The asterisk
    indicates that while there were actually 3 thresholds, since they
    occur in 3:2:1 ratio, they doubled the number of languages.}
  \label{tab:lgs2}
\end{table}

We recognize that the composition of MLRegTest is unbalanced. As
Tables~\ref{tab:lgs1} and~\ref{tab:lgs2} show, some classes have more
languages than others. The largest disparity is between TLTT with 300
languages, and SF, $\mathbb{Z}_p$ and Reg, which each have 30
languages. Nonetheless, what matters for the experimental design and
statistical analysis is that each class contains a representative
sample of languages, and each class in MLRegTest contains at least 30
languages. That the statistical analysis itself is not weakened by
these disparities is discussed in some detail in the discussion
in~\S\ref{ExperimentalDesign} of the experimental design and
analytical techniques. While we cannot guarantee that these 30 are the
most representative languages in the class, we believe they are more
representative, as a whole, of these classes than those in previous
research.

An automaton representing each language was generated by the Language
Toolkit (LTK) \citep{Lambert2024} from files readable by LTK.  Those
files were generated by a Python program.\footnote{In the software,
  languages were named according to the scheme
  \textsc{sigma.tau.class.k.t.i}, where \textsc{sigma} is a two-digit
  alphabet size, \textsc{tau} a two-digit number of salient symbols
  (the `tier'), \textsc{class} the named subregular class, \textsc{k}
  the width of factors used (if applicable), \textsc{t} the threshold
  counted to (if applicable), and \textsc{i} a unique identifier.} The
Language Toolkit extends traditional regular expressions with basic
terms that are interpreted as languages which contain substrings
and/or subsequences. This does not increase the expressivity of
traditional regular expressions, but it does facilitate the
construction of languages belonging to the aforementioned
classes. These expressions are then compiled into finite-state
automata.

The languages in the coSL, TcoSL, and coSP classes were chosen to be
the complements of the languages in the SL, TSL, and SP classes,
respectively.

For each class $C$ above, the programs The Language Toolkit and
Amalgam include algorithms which decide whether a given finite-state
automaton belongs to $C$. Therefore, to verify that a language $L$
counts as a representative of class $C_0$ and not to classes $C_1$,
$C_2$ and so on, we ran the decision algorithms for classes
$C_0, C_1, C_2$ on the finite-state automaton for $L$ and ensured that
$L$ belonged to $C$ but not to $C_1, C_2$ and so on. This was done for
each of the 1,800 languages in MLRegTest. 

The decision procedures for many of these classes can be found in the
algebraic literature on automata theory \citep{Pin2021}. The decision
procedures for the tier-based classes are presented in
\citep{Lambert2023}. These procedures take as input either the minimal
DFA corresponding to the language or the syntactic monoid
corresponding to the language. The minimal DFA for a language \(L\) is
the acceptor whose states correspond to the blocks of the coarsest
partition of \(L\) that forms a right congruence \citep[the Nerode
relation]{ScottRabin1959}. The syntactic monoid for a language
\(L\) is the acceptor whose states correspond to the blocks of the
coarsest partition of \(L\) that forms a congruence \citep[the Myhill
relation]{ScottRabin1959}. The Myhill relation refines the Nerode
relation, and computing the syntactic monoid from the minimal DFA is
in the worst case exponential. Nonetheless, we were able to construct
the syntactic monoids for all the languages in MLRegTest using The
Language Toolkit.  Generally speaking, the decision procedures run in
time polynomial in the size of the syntactic monoid or in the size of
the minimal DFA.

Amalgam typically consumes considerably less time and memory in
practice than The Language Toolkit when deciding class
membership. Using Amalgam, we verified that every language $L$ labeled
as belonging to class $C$ in MLRegTest (except those belonging to the
SP and coSP classes) counts as a representative of class $C$. The only
exceptions to this are the Strictly Piecewise class and its
complement, for which Amalgam currently has no test. For these
classes, using The Language Toolkit, we were able to verify that all
the languages in SP and coSP count as a representative of class SP and
coSP, respectively. In this way, all languages in MLRegTest were
verified as being representative of their designated class.

\subsection{Randomly Constructing Finite-State Automata}

One motivation for the careful curation and construction of the
languages in MLRegTest was that languages in most of these classes are
unlikely to be generated randomly using straightforward procedures.
As evidence for this claim, we randomly constructed finite-state
automata of different sizes with two parameters, one controlling the
probability a state was accepting, and one controlling whether a
transition existed between two states.  We then used some of the
decision procedures mentioned above to classify them, and found they
mostly belonged to the SL class.

The procedure we used for randomly generating machines was as
follows. We fixed a number of states $n$, a number of symbols $s$, a
start state, an edge-probability $0\leq p_e\leq 1$, and an
acceptance-probability $0\leq p_f\leq 1$.  For each state, it was
accepting with probability $p_f$.  For each $\sigma\in\Sigma$, and for
each pair of states $(q,r)$, we included an edge from $q$ to $r$
labeled $\sigma$ with probability $p_e$.  We ran several experiments,
varying in each of these parameters.  Our primary result is that as
$n$ increases, it was more likely that the automaton generated was
Strictly Local.

\begin{figure}\centering
  \input{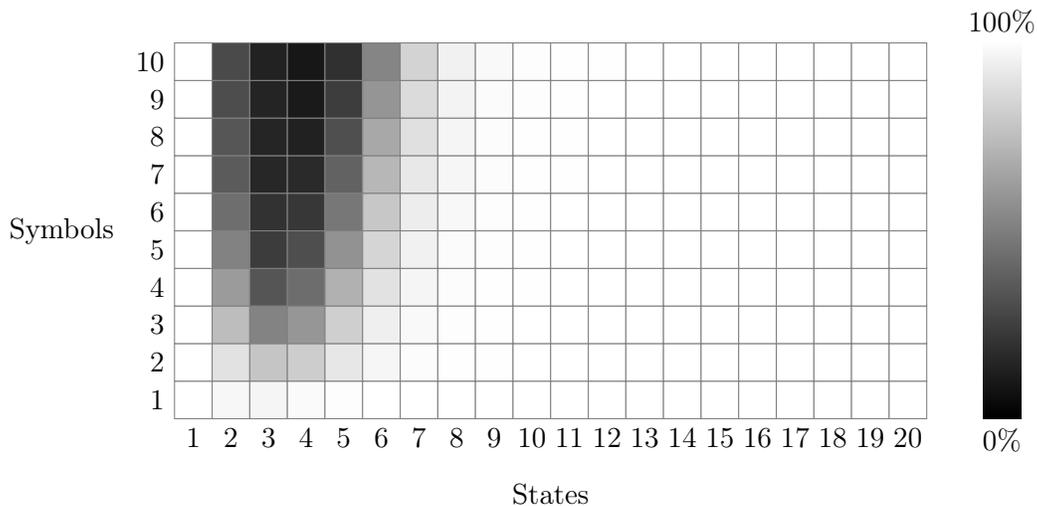}
  \caption{The proportion of Strictly Local languages
    upon fair generation, $p_e=p_f=0.5$.}
  \label{fig:sl-5050}
\end{figure}

We began with a fair construction, with $p_e=p_f=0.5$, varying $q$ and
$s$ over a range of values.  For each combination of $q\in[1,20]$ and
$s\in[1,10]$, we generated ten thousand automata by this method and
determined how many of those ten thousand were Strictly Local.  A heat
map of the results is shown in Figure~\ref{fig:sl-5050}.  As one can
see, unless the alphabet is sufficiently large compared to the number
of states, a vast majority of languages generated by this method are
Strictly Local.  The mean average had 87.57\% in this class, with
$n=7$ and $s=8$ as the parameters yielding the result closest to this
value.

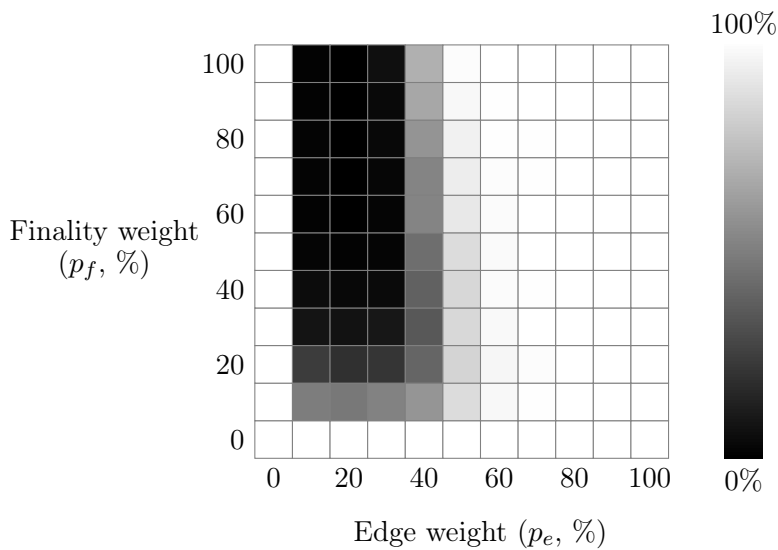
\begin{figure}\centering
  \begin{tikzpicture}[x=5mm,y=5mm]
\draw[gray,fill=black!0!white] (-1,-1) rectangle (0,0);
\draw[gray,fill=black!0!white] (-1,0) rectangle (0,1);
\draw[gray,fill=black!0!white] (-1,1) rectangle (0,2);
\draw[gray,fill=black!0!white] (-1,2) rectangle (0,3);
\draw[gray,fill=black!0!white] (-1,3) rectangle (0,4);
\draw[gray,fill=black!0!white] (-1,4) rectangle (0,5);
\draw[gray,fill=black!0!white] (-1,5) rectangle (0,6);
\draw[gray,fill=black!0!white] (-1,6) rectangle (0,7);
\draw[gray,fill=black!0!white] (-1,7) rectangle (0,8);
\draw[gray,fill=black!0!white] (-1,8) rectangle (0,9);
\draw[gray,fill=black!0!white] (-1,9) rectangle (0,10);
\draw[gray,fill=black!0!white] (0,-1) rectangle (1,0);
\draw[gray,fill=black!50.9!white] (0,0) rectangle (1,1);
\draw[gray,fill=black!76.5!white] (0,1) rectangle (1,2);
\draw[gray,fill=black!92.1!white] (0,2) rectangle (1,3);
\draw[gray,fill=black!95.1!white] (0,3) rectangle (1,4);
\draw[gray,fill=black!98.1!white] (0,4) rectangle (1,5);
\draw[gray,fill=black!98.9!white] (0,5) rectangle (1,6);
\draw[gray,fill=black!98.6!white] (0,6) rectangle (1,7);
\draw[gray,fill=black!98.5!white] (0,7) rectangle (1,8);
\draw[gray,fill=black!98.9!white] (0,8) rectangle (1,9);
\draw[gray,fill=black!99!white] (0,9) rectangle (1,10);
\draw[gray,fill=black!0!white] (1,-1) rectangle (2,0);
\draw[gray,fill=black!52.8!white] (1,0) rectangle (2,1);
\draw[gray,fill=black!81!white] (1,1) rectangle (2,2);
\draw[gray,fill=black!92.8!white] (1,2) rectangle (2,3);
\draw[gray,fill=black!96.7!white] (1,3) rectangle (2,4);
\draw[gray,fill=black!98.8!white] (1,4) rectangle (2,5);
\draw[gray,fill=black!100!white] (1,5) rectangle (2,6);
\draw[gray,fill=black!100!white] (1,6) rectangle (2,7);
\draw[gray,fill=black!100!white] (1,7) rectangle (2,8);
\draw[gray,fill=black!99.9!white] (1,8) rectangle (2,9);
\draw[gray,fill=black!100!white] (1,9) rectangle (2,10);
\draw[gray,fill=black!0!white] (2,-1) rectangle (3,0);
\draw[gray,fill=black!49.1!white] (2,0) rectangle (3,1);
\draw[gray,fill=black!79.1!white] (2,1) rectangle (3,2);
\draw[gray,fill=black!91.1!white] (2,2) rectangle (3,3);
\draw[gray,fill=black!96.1!white] (2,3) rectangle (3,4);
\draw[gray,fill=black!97.9!white] (2,4) rectangle (3,5);
\draw[gray,fill=black!98.2!white] (2,5) rectangle (3,6);
\draw[gray,fill=black!97.7!white] (2,6) rectangle (3,7);
\draw[gray,fill=black!96.7!white] (2,7) rectangle (3,8);
\draw[gray,fill=black!96.3!white] (2,8) rectangle (3,9);
\draw[gray,fill=black!94!white] (2,9) rectangle (3,10);
\draw[gray,fill=black!0!white] (3,-1) rectangle (4,0);
\draw[gray,fill=black!41.7!white] (3,0) rectangle (4,1);
\draw[gray,fill=black!60.2!white] (3,1) rectangle (4,2);
\draw[gray,fill=black!65.4!white] (3,2) rectangle (4,3);
\draw[gray,fill=black!61.8!white] (3,3) rectangle (4,4);
\draw[gray,fill=black!57!white] (3,4) rectangle (4,5);
\draw[gray,fill=black!48.3!white] (3,5) rectangle (4,6);
\draw[gray,fill=black!48.3!white] (3,6) rectangle (4,7);
\draw[gray,fill=black!42!white] (3,7) rectangle (4,8);
\draw[gray,fill=black!34.6!white] (3,8) rectangle (4,9);
\draw[gray,fill=black!31!white] (3,9) rectangle (4,10);
\draw[gray,fill=black!0!white] (4,-1) rectangle (5,0);
\draw[gray,fill=black!13.6!white] (4,0) rectangle (5,1);
\draw[gray,fill=black!16.9!white] (4,1) rectangle (5,2);
\draw[gray,fill=black!15.4!white] (4,2) rectangle (5,3);
\draw[gray,fill=black!15.7!white] (4,3) rectangle (5,4);
\draw[gray,fill=black!13.7!white] (4,4) rectangle (5,5);
\draw[gray,fill=black!9.1!white] (4,5) rectangle (5,6);
\draw[gray,fill=black!6.9!white] (4,6) rectangle (5,7);
\draw[gray,fill=black!5.3!white] (4,7) rectangle (5,8);
\draw[gray,fill=black!2.9!white] (4,8) rectangle (5,9);
\draw[gray,fill=black!1.1!white] (4,9) rectangle (5,10);
\draw[gray,fill=black!0!white] (5,-1) rectangle (6,0);
\draw[gray,fill=black!2.9!white] (5,0) rectangle (6,1);
\draw[gray,fill=black!3.7!white] (5,1) rectangle (6,2);
\draw[gray,fill=black!2.5!white] (5,2) rectangle (6,3);
\draw[gray,fill=black!1.7!white] (5,3) rectangle (6,4);
\draw[gray,fill=black!2.1!white] (5,4) rectangle (6,5);
\draw[gray,fill=black!1.6!white] (5,5) rectangle (6,6);
\draw[gray,fill=black!1.1!white] (5,6) rectangle (6,7);
\draw[gray,fill=black!0.4!white] (5,7) rectangle (6,8);
\draw[gray,fill=black!0.2!white] (5,8) rectangle (6,9);
\draw[gray,fill=black!0!white] (5,9) rectangle (6,10);
\draw[gray,fill=black!0!white] (6,-1) rectangle (7,0);
\draw[gray,fill=black!0.3!white] (6,0) rectangle (7,1);
\draw[gray,fill=black!1.1!white] (6,1) rectangle (7,2);
\draw[gray,fill=black!0!white] (6,2) rectangle (7,3);
\draw[gray,fill=black!0.3!white] (6,3) rectangle (7,4);
\draw[gray,fill=black!0.1!white] (6,4) rectangle (7,5);
\draw[gray,fill=black!0.1!white] (6,5) rectangle (7,6);
\draw[gray,fill=black!0.2!white] (6,6) rectangle (7,7);
\draw[gray,fill=black!0.4!white] (6,7) rectangle (7,8);
\draw[gray,fill=black!0!white] (6,8) rectangle (7,9);
\draw[gray,fill=black!0!white] (6,9) rectangle (7,10);
\draw[gray,fill=black!0!white] (7,-1) rectangle (8,0);
\draw[gray,fill=black!0!white] (7,0) rectangle (8,1);
\draw[gray,fill=black!0.1!white] (7,1) rectangle (8,2);
\draw[gray,fill=black!0.1!white] (7,2) rectangle (8,3);
\draw[gray,fill=black!0!white] (7,3) rectangle (8,4);
\draw[gray,fill=black!0.1!white] (7,4) rectangle (8,5);
\draw[gray,fill=black!0.1!white] (7,5) rectangle (8,6);
\draw[gray,fill=black!0!white] (7,6) rectangle (8,7);
\draw[gray,fill=black!0!white] (7,7) rectangle (8,8);
\draw[gray,fill=black!0!white] (7,8) rectangle (8,9);
\draw[gray,fill=black!0!white] (7,9) rectangle (8,10);
\draw[gray,fill=black!0!white] (8,-1) rectangle (9,0);
\draw[gray,fill=black!0!white] (8,0) rectangle (9,1);
\draw[gray,fill=black!0!white] (8,1) rectangle (9,2);
\draw[gray,fill=black!0!white] (8,2) rectangle (9,3);
\draw[gray,fill=black!0!white] (8,3) rectangle (9,4);
\draw[gray,fill=black!0!white] (8,4) rectangle (9,5);
\draw[gray,fill=black!0!white] (8,5) rectangle (9,6);
\draw[gray,fill=black!0!white] (8,6) rectangle (9,7);
\draw[gray,fill=black!0!white] (8,7) rectangle (9,8);
\draw[gray,fill=black!0!white] (8,8) rectangle (9,9);
\draw[gray,fill=black!0!white] (8,9) rectangle (9,10);
\draw[gray,fill=black!0!white] (9,-1) rectangle (10,0);
\draw[gray,fill=black!0!white] (9,0) rectangle (10,1);
\draw[gray,fill=black!0!white] (9,1) rectangle (10,2);
\draw[gray,fill=black!0!white] (9,2) rectangle (10,3);
\draw[gray,fill=black!0!white] (9,3) rectangle (10,4);
\draw[gray,fill=black!0!white] (9,4) rectangle (10,5);
\draw[gray,fill=black!0!white] (9,5) rectangle (10,6);
\draw[gray,fill=black!0!white] (9,6) rectangle (10,7);
\draw[gray,fill=black!0!white] (9,7) rectangle (10,8);
\draw[gray,fill=black!0!white] (9,8) rectangle (10,9);
\draw[gray,fill=black!0!white] (9,9) rectangle (10,10);
\foreach \x in {0,2,...,10} {
\node[anchor=north] at (\x-0.5,-1) {\if\x0\x\else\x0\fi};
}
\node at (5,-3) {Edge weight ($p_e$, \%)};
\foreach \y in {0,2,...,10} {
\node[anchor=east] at (-1,\y-0.5) {\if\y0\y\else\y0\fi};
}
\node at (-5,5) (pf) {Finality weight};
\node[anchor=north] at (pf.base) {($p_f$, \%)};

\node[anchor=north] at (12,-1) {0\%};
\node[anchor=south] at (12,10) {100\%};
\fill[left color=white,right color=black,shading=axis,shading angle=0] (11.5,-1) rectangle (12.5,10);
\end{tikzpicture}
  \caption{The proportion of Strictly Local languages for 7 states, 8 symbols.}
  \label{fig:sl-7q8a}
\end{figure}

From there, we fixed $n=7$ and $s=8$ and varied $p_e$ and $p_f$
from $0$ to $1$ in intervals of $0.1$.
For each parameterization here, we generated one thousand machines
and cataloged which were Strictly Local.
Of course, when the $p_f$ is exactly $0$ or exactly $1$,
the resulting language is the empty set or its complement,
and thus strictly local,
so those cases are not exactly interesting.
And if $p_e$ is exactly $0$ only the empty set is generated.
But outside of these special cases,
the effect of $p_f$ is dwarfed by that of $p_e$,
where a sparser graph is significantly less likely to be Strictly Local.
The heat map is shown in Figure~\ref{fig:sl-7q8a}.

In sum, we cannot in good faith recommend random generation as a
mechanism for producing test languages, as, without careful
consideration of parameterization, the resulting languages are
overwhelmingly Strictly Local.  As this is among the simplest possible
subregular classes, such generation could easily lead one to believe
that a machine-learning algorithm performs significantly better than
it might on a more diverse set of regular languages.




\section{Data Sets} 
\label{sec:ds}
For each language in MLRegTest, we separately generated training,
development, and test data sets. To generate data sets, we used the
software library Pynini \citep{Gorman2016,GormanSproat2021}, which is
a Python front-end to OpenFst \citep{openfst}.\footnote{OpenFst is
available at \url{https://www.openfst.org/twiki/bin/view/FST/WebHome}
and Pynini at
\url{https://www.openfst.org/twiki/bin/view/GRM/Pynini}.} The
automaton constructed with The Language Toolkit was exported to the
\texttt{att} format and then converted to a binary format by OpenFst,
which is a format Pynini reads.

For each language $L$ we generated a training set which included
100,000 strings, half of which belonged to the language and half of
which did not. We call the strings belonging to $L$ positive, and the
strings not belonging to $L$ negative.

We generated equally many strings of length $\ell$ where $\ell$ ranged
between 20 and 29. We chose a minimum length of 20 to ensure that we
could generate enough positive and negative strings for each
language. Some language have a very few positive or negative strings
at shorter lengths. As an extreme example, the shortest negative
string in language 64.11.TLTT.4.3.1 is of length 12. We note that
length 20 may not be the minimum length we could have used, but we
found it was sufficient.

The positive and negative strings in the training sets were generated
in a few steps. For the positive strings, the automaton for $L$ was
first intersected with the automaton for $\Sigma^\ell$. Second,
probabilities were assigned to the edges of this acyclic automaton to
ensure a uniform distribution over its paths. This was accomplished
with a reverse topological sort. Finally, paths were selected randomly
from this weighted automaton.\footnote{We observed that randomly
selecting paths by assuming a uniform distribution on outbound edges
was not effective. For example, consider the coSL language where words
must contain an $aa$ substring (language 64.64.coSL.2.1.0). If there
is a uniform distribution over outbound edges in the acyclic automaton
generating words of this language of length 20, then in the first
state, the probability of selecting the edge labeled $a$ is
1/64. Similarly, for the second state. In general, the probability of
producing an $aa$ substring at any given point is $(1/64)^2=1/4096$.
It is very unlikely for $aa$ to occur by chance under these
conditions, for all but the latest states. If no $aa$ substring has
occurred by the antepenultimate state, then the probability of
selecting an edge with $a$ becomes 1. And similarly for the
penultimate state for the simple reason that the this acyclic machine
only generates words with the substring $aa$. Assuming uniform
distribution over the outbound edges of each state for this language
has the consequence that $aa$ substrings overwhelmingly occur at the
right edge of the word.} The negative strings were similarly generated
using the complement of $L$. The net effect of these decisions is that
all positive (or negative) examples of a given length are equally
likely to be chosen. Note that it was possible for the same string to
be generated more than once (duplicates were allowed).

For each language $L$ we similarly generated a development set which
included 100,000 strings, half of which were positive and half of
which were negative. As with the training set, there were equally many
strings of length $\ell$ where $\ell$ ranged between 20 and 29. The
positive strings were generated by intersecting the automaton for $L$
with the automaton for $\Sigma^\ell$, removing the positive strings
from the training set, and then weighting the edges of this acyclic
automaton as before to ensure a uniform distribution over the
paths. The negative strings were similarly generated using the
complement of $L$ and removing the negative strings in the training
set. In this way, we ensured the training and development sets for
every $L$ were disjoint.

For each language $L$ we generated four test sets, each with 100,000
strings, half of which were positive and half of which were
negative. We call these four test sets ``Short Random'' (SR), ``Short
Adversarial'' (SA), ``Long Random'' (LR), ``Long Adversarial''
(LA). The Short Test sets included equally many strings of length
$\ell$ where $\ell$ ranged between 20 and 29. The Long Test sets
included equally many strings of length $\ell$ where $\ell$ ranged
between 31 and 50. The Random Test sets sampled positive and negative
strings without replacement. In the Adversarial Test sets, each
positive string $x$ was paired with a negative string $y$ such that
the string edit distance $d(x,y)=1$. No positive or negative string in
any Test set occurred in the Training or Development sets. Below we
describe how we generated the data to meet these specifications.

The Short Random Test sets generated positive strings as follows. For
each length $\ell$, the automaton $A$ was constructed by intersecting
the automaton for $L$ with the automaton for $\Sigma^\ell$, and
removing the positive strings from both the training and development
sets. The acyclic automaton $A$ was weighted to ensure a uniform
probability distribution over its paths. Then the following procedure
was repeated. Let $n$ be the number of strings remaining to be
generated (initially 5,000) and $P$ the list of strings currently
obtained (initially empty). Then $n$ many positive strings were
generated by selecting $n$ paths from $A$. Strings were added to a
list only if they did not already occur in this list. Then $n$ was
updated to $5,000$ minus the length of this accumulating list.
This process repeated until all desired unique strings were obtained.
A similar procedure was followed for generating the negative strings.
In this way, we ensured the SR Test set was
disjoint from both the training and development sets, and that each
string in the SR Test sets was unique. The Long Random Test sets were
generated similarly by randomly sampling strings of each length
without replacement.

The Short Adversarial Test sets for each $L$ were constructed
according to the following procedure. We constructed the transducer
$C\circ T\circ A$ where $A$ is the original automaton used to
construct the SR Test, $T$ is the transducer recognizing the relation $\{
(x,y) \mid x,y\in\Sigma^*, d(x,y)=1\}$, and $C$ is the automaton
recognizing the complement of $L$, and where $\circ$ indicates
composition. Consequently $C\circ T\circ A$ is the transducer whose
paths correspond to positive strings $x$ of length $\ell$ and negative
strings $y$ such that $d(x,y)=1$. This machine was weighted to ensure
a uniform distribution over its paths. For each $\ell$, 5,000 unique
paths were randomly selected to ultimately obtain 50,000 unique pairs
of positive and negative strings. The Long Adversarial Test sets were
generated similarly to the SA Test sets.

The above procedures produced 6 data sets (Train, Dev, SR, SA, LR,
LA), each with 50,000 positive and 50,000 negative strings.  We then
made additional Train, Dev, SR, SA, LR, LA sets of 1/10th and 1/100th
the size by downsampling. Consequently, for every language we prepared
3 training sets, 3 development sets and 12 test sets. The sets with
100,000 words we call ``Large'', those with 10k words we call
``Mid'', and those with 1,000 words we call ``Small.'' These sets
are nested so that every string in the Small set is included in the
Mid set, which is included in Large set.

The above procedures were followed for all languages except the
languages in the coSL, TcoSL, and coSP classes. The datasets for coSL,
TcoSL, and coSP languages were generated simply by switching the
positive and negative strings in the corresponding datasets for the
corresponding SL, TSL and SP languages.









\section{Experiments} 
This section reports on the experiments that were conducted to assess
the capabilities of generic neural networks to model the languages in
MLRegTest. Our goal is to obtain a fine-grained understanding of the
strengths and limitations of neural networks in modeling regular
languages. We analyze the associations between neural network
performance and the linguistic and model parameters listed in
\Cref{variables-table}. By independently training and evaluating
neural networks on nearly all combinations of the factors in
\Cref{variables-table}, a large sample size ($n=86,400$) of accuracy
scores was collected, making possible powerful tests of statistical
association.

Four neural network architectures---simple recurrent neural network
(RNN), gated recurrent unit (GRU), long short-term memory (LSTM), and
transformer---were employed in our experiments. One of the
considerations shaping our experimental design is that we are not
analyzing the ways in which model hyperparameters such as learning
rate, embedding dimension, and loss function correlate with model
performance across MLRegTest. Accordingly, a fixed set of
hyperparameters was obtained for each neural network architecture via
a preliminary hyperparameter search, described in
\S\ref{subsec:ExpHyperparams}. The main experiments, using the fixed
hyperparameters from the preliminary step, follow a factorial design
described in \S\ref{ExperimentalDesign}.

Throughout our analysis of the experimental results, we use accuracy
as the response variable. We justify the use of accuracy since
positive and negative data are balanced in MLRegTest, ensuring that
there is no bias implicit in the dataset. Further, alternative
measures of neural network performance including F-score, precision,
and Brier score each correlate strongly with accuracy
(\Cref{corrs-table}).
\begin{table}[h]
  \centering
  \begin{tabular}{lrrr}
    \toprule
            & \multicolumn{1}{c}{Accuracy} & \multicolumn{1}{c}{AUC} & \multicolumn{1}{c}{Brier} \\
    \midrule
    AUC     & \(0.977\)                & \centerdash             & \centerdash               \\
    Brier   & \(-0.970\)               & \(-0.950\)          & \centerdash               \\
    F-score & \(0.882\)                & \(0.867\)           & \(-0.845\)            \\
    \bottomrule
  \end{tabular}
  \caption{Correlation matrix of performance metrics.}
  \label{corrs-table}
\end{table}

\subsection{Neural Network Details}
The Tensorflow \citep{tensorflow2015-whitepaper} and Keras
\citep{chollet2015keras} APIs were used throughout the
experiments. Each neural network consisted of the following ordered
modules: trainable embedding (with random initialization and embedding dimension of 32 or 256, as determined by grid
search and described in \S\ref{subsec:ExpHyperparams}); unidirectional recurrent module (simple
RNN, GRU, or LSTM) or two consecutive transformer blocks (each with
two attention heads, dropout of 0.2, and layer norm epsilon parameter of \num{1e-6}); dense feed-forward layers
(each with output dimension 64, and number of layers chosen by grid
search); dropout (chosen by grid search); layer normalization ($\epsilon=\num{1e-6}$); and
softmax activation. The number of hidden states in the RNNs and the dimensionality of the key vectors in the
transformer architecture were equal to the embedding dimension mentioned above. All neural
networks were trained with a batch size of 64 and used binary
cross-entropy (BCE) loss.

\subsection{Hyperparameter Search}\label{subsec:ExpHyperparams}
For each of the four architecture types, we obtained a fixed set of
hyperparameters to use throughout the main experiments. The search was
organized as follows. A representative selection of 32 languages from
MLRegTest was chosen according to the following criteria: two
languages from each of the sixteen language classes; all with alphabet
size sixteen; factor width 0, 3, or 4, whichever applies; threshold
value 0, 1, or 2, whichever applies; and identification numbers 3 and
6. These 32 languages are listed in \Cref{tab:search-lgs} in the appendix. These values were
chosen because they were either the only value (0) or the middle value
in the options available for those languages.

For all architecture types and all languages in the selection, we ran
an exhaustive search over all models in the following hypergrid:
number of feed-forward layers (2 or 4); embedding dimension (32 or
256); learning rate (0.01 or 0.0001); dropout (0.0 or 0.1); number of
epochs (32 or 64); loss function (binary cross-entropy or mean squared
error); and optimizer (RMSProp, Adam, or SGD). For every model in the
hyperparameter search, we used the Medium sized training and
validation sets.  The validation set for the corresponding language
was split in half: one half was used for validation during training
and the other half was used for evaluating the accuracy of the
model. The validation sets were used in the grid search for testing
(as opposed to the test sets) to avoid statistical bias in the main
experiments.

Hyperparameters were selected from the grid search as follows. For
each architecture type and for each setting of the hyperparameters (of
which there were $2^6\cdot3$ total settings), we computed the average
accuracy over all 32 languages. The hyperparameter setting with the
greatest of those mean accuracies was selected for that architecture
type. In all cases, the greatest mean accuracy was unique. The results
of the grid search are listed in \Cref{gridsearch-table}.

\begin{table}
  \centering
  \begin{tabular}{l *{5}{c}}
    \toprule
                                  & \multicolumn{4}{c}{Network Type}             \\ \cline{2-5}
                                  &                                              \\[-9pt]
    Hyperparameters               & Simple RNN & GRU     & LSTM    & Transformer \\ 
    \midrule
    Learning Rate                 & 0.0001     & 0.01    & 0.0001  & 0.0001      \\
    Optimizer                     & Adam       & RMSProp & RMSProp & Adam        \\
    Number of Epochs              & 64         & 64      & 64      & 64          \\
    Loss Function                 & BCE        & BCE     & BCE     & BCE         \\
    Embedding Dimension           & 32         & 32      & 256     & 256         \\
    Number of Feed Forward Layers & 4          & 2       & 2       & 2           \\
    Dropout                       & 0.1        & 0.1     & 0.0     & 0.1         \\
    \bottomrule
  \end{tabular}
  \caption{Hyperparameters selected from grid search.}
  \label{gridsearch-table}
\end{table}

Fixing the hyperparameters given in \Cref{gridsearch-table}, the
number of trainable parameters of the neural networks is listed by
alphabet size and network type in \Cref{params-table}. The notable
difference in number of trainable parameters between network types
stems from the different embedding dimensions, namely 32 versus 256.
\begin{table}
  \centering
  \begin{tabular}{l *{5}{c}}
    \toprule
                  & \multicolumn{4}{c}{Network Type}             \\ \cline{2-5}
                  &                                              \\[-9pt]
    Alphabet Size & Simple RNN & GRU     & LSTM    & Transformer \\ 
    \midrule
    4             & 17,090     & 13,026  & 547,458 & 1,091,714   \\
    16            & 17,474     & 13,410  & 550,530 & 1,094,786   \\
    64            & 19,010     & 14,946  & 562,818 & 1,107,074   \\
    \bottomrule
  \end{tabular}
  \caption{Number of trainable parameters by alphabet size and network type.}
  \label{params-table}
\end{table}

\subsection{Experimental Design}\label{ExperimentalDesign}
This subsection describes the design of our main set of
experiments. All neural networks in these experiments use the
hyperparameters determined by the grid search described in
\S\ref{subsec:ExpHyperparams}. We examine the effects of the design
factors in \Cref{variables-table} on model accuracy: six factors
describe languages, two describe datasets, and one describes neural
network architecture. A realization of the factors \Alpha{}, \Tier{},
\Class{}, \kay{}, \jay{}, and \eye{} specifies a regular language in
MLRegTest (see \S\ref{subsec:mlregtest} for details). A \textit{model
  configuration} in the present context refers to a choice of regular
language, training set size, and neural network architecture, which is
to say a realization of all factors in \Cref{variables-table} except
\TestType{}. Our experimental design consists of
\begin{center}
  1,800 (languages) $\times$ 3 (training set sizes) $\times$ 4 (NN
  architectures) $=$ 21,600
\end{center}
model configurations, each corresponding to a unique trained
model. Models were trained with some \TrainSize{}. The correspondingly
sized validation set was solely used to monitor (without any
intervention) the progress of the networks during training. For all
model configurations in the design, the associated trained model was
evaluated on the four distinct test sets SR, SA, LR, and LA (described
in \S\ref{sec:ds}) corresponding with the model configuration's
regular language. In total, 86,400 observations of model accuracy
scores were collected.
\begin{table}[h]
    \centering
    \begin{tabular}{l p{10cm}}
        \toprule
        Factor Name & Description (Levels)                                                                                     \\
        \midrule
        \Alpha      & Alphabet size (4, 16, 64)                                                                                \\
        \Tier       & (2, 3, 4, 6, 7, 11, 16, 64)                                                                              \\
        \Class      & Subregular language class (SL, coSL, TSL, TcoSL, SP, coSP, LT, TLT, PT, LTT, TLTT, PLT, TPLT, SF, Zp, Reg) \\
        \kay        & Factor width (0, 2, 3, 4, 5, 6)                                                                          \\
        \jay        & Threshold (0, 1, 2, 3, 5)                                                                                \\
        \eye        & Language identification number (0, 1, 2, 3, 4, 5, 6, 7, 8, 9)                                            \\
        \TrainSize  & Size of training set (Small: 1k, Mid: 10k, Large: 100k)                                                  \\
        \NNType     & Neural network architecture (Simple RNN, GRU, LSTM, Transformer)                           \\
        \TestType   & Type of test set: whether strings are short or long, random or adversarial (SR, SA, LR, LA)              \\
        \bottomrule
    \end{tabular}
    \caption{Factors comprising the experimental design.}
    \label{variables-table}
\end{table}

All 2,304 combinations of \Alpha{}, \Class{}, \TrainSize{},
\NNType{}, and \TestType{} were tested in our
experiments. Importantly, not all combinations of \Tier{}, \kay{},
\jay{}, and \eye{} were tested because the values of these parameters
depend on those of \Alpha{} and \Class{}. For example, not all
language classes have different tier alphabets or thresholds (see
\S\ref{sec:lgs}).

Our experimental design and statistical analysis follows the
approaches advocated by \citet{Demsar2006} and \citet{Stapor2017},
namely the use of multiple comparisons statistical tests to compare
the performance of learning algorithms. We form a full factorial
design by mean-aggregating the factors \Tier{}, \kay{}, \jay{}, and
\eye{}, that is, the value of each cell of the design is the average
of accuracy scores of observations that have the same values of
\Alpha{}, \Class{}, \TrainSize{}, \NNType{}, and \TestType{} but
different values of \Tier{}, \kay{}, \jay{}, and \eye{}. There are
thus 2,304 cells in the design, indexed by the factors \Alpha{},
\Class{}, \TrainSize{}, \NNType{}, and \TestType{}. This design admits
repeated-measures non-parametric statistical tests, in particular the
Friedman test \citep{Demsar2006,Stapor2017}: any of the five factors
\Alpha{}, \Class{}, \TrainSize{}, \TestType{}, or \NNType{} can be
considered a treatment variable (i.e. we hypothesize that these
factors may be associated with accuracy), while the remaining four
form blocking variables.

To apply the Friedman test, the full factorial design is cast to a
matrix whose rows correspond with distinct combinations of blocking
variable levels and whose columns represent treatment levels. The
Friedman test is then invoked to ask whether any of the treatment
levels differ. In those cases that we reject the Friedman test null
hypothesis of equal treatment effects, we conduct post hoc analyses
using the Nemenyi--Wilcoxon--Wilcox all-pairs test to determine
precisely which treatment levels are pairwise significantly different
from one another \citep{Pereira+2015,Singh+2016}. Using this general
framework for statistical analysis of the experimental design, we
obtain a fine-grained understanding of how the design factors---as
well as factors like logical level and order relation, which are
inferred from language classes---associate with accuracy. The results
of this analysis are presented in \S\ref{sec:res}.

Furthermore, this statistical analysis is affected only minimally by
the disparities existing among language classes (see discussion around
Tables~\ref{tab:lgs1} and~\ref{tab:lgs2}). The design only sees
alphabet size and language class, so the structure of the design is no
different than if the classes were balanced (had equally many
languages per class). The disparity does imply that different cells of
the design have different variances: classes with more languages yield
cells with lower variance. Lower variances are only welcome. The key
point is that since each class has a representative sample, the
difference in variances is mostly inconsequential.

Finally, the Friedman test and Nemenyi--Wilcoxon--Wilcox all-pairs
test both report p-values, which represent ``the probability,
calculated under the null hypothesis, that a test statistic is as
extreme or more extreme than its observed value''
\citep{Benjamin2018}. If this probability is deemed low enough to be
significant, then the null hypothesis is rejected and it is concluded
that an effect is present. It is therefore important to determine the
cutoff $\alpha$, known as the significance level, at which p-values
are deemed significant. Many communities set the significance level to
$\alpha=0.05$ though \citet{Benjamin2018} argue it should be set lower
to $\alpha=0.005$. Some communities, such as researchers in
high-energy physics and genetics, have stricter levels. We follow the
recommendation of \citet{Benjamin2018}, though we note that nearly all
of our results of statistical significance are established by p-values
on the order of $10^{-4}$ or smaller.

\section{Results} 
\label{sec:res}
This section presents the results obtained from the experimental
design described in \Cref{ExperimentalDesign} as well as their
interpretation with regards to salient research
questions.\footnote{The results are collected in the file
  \texttt{all\_evals.csv} which is located in the ``analysis''
  directory in the project's github repository
  \url{https://github.com/heinz-jeffrey/subregular-learning}. The
  results presented here were processed with the \texttt{analysis.R}
  file, also located in the directory.}

\subsection{Sanity Checks}
\label{sec:sanity}

We first report some results that give us confidence in the validity
of our experimental setup and factorial design.

\subsubsection{Training Size}
Because the Small, Mid, and Large data sets are nested, we fully
expect additional data will improve accuracy. Our first question was
whether our results bore out this expectation.

\begin{figure}[ht]
  \centering
  \includegraphics[width=1\textwidth]{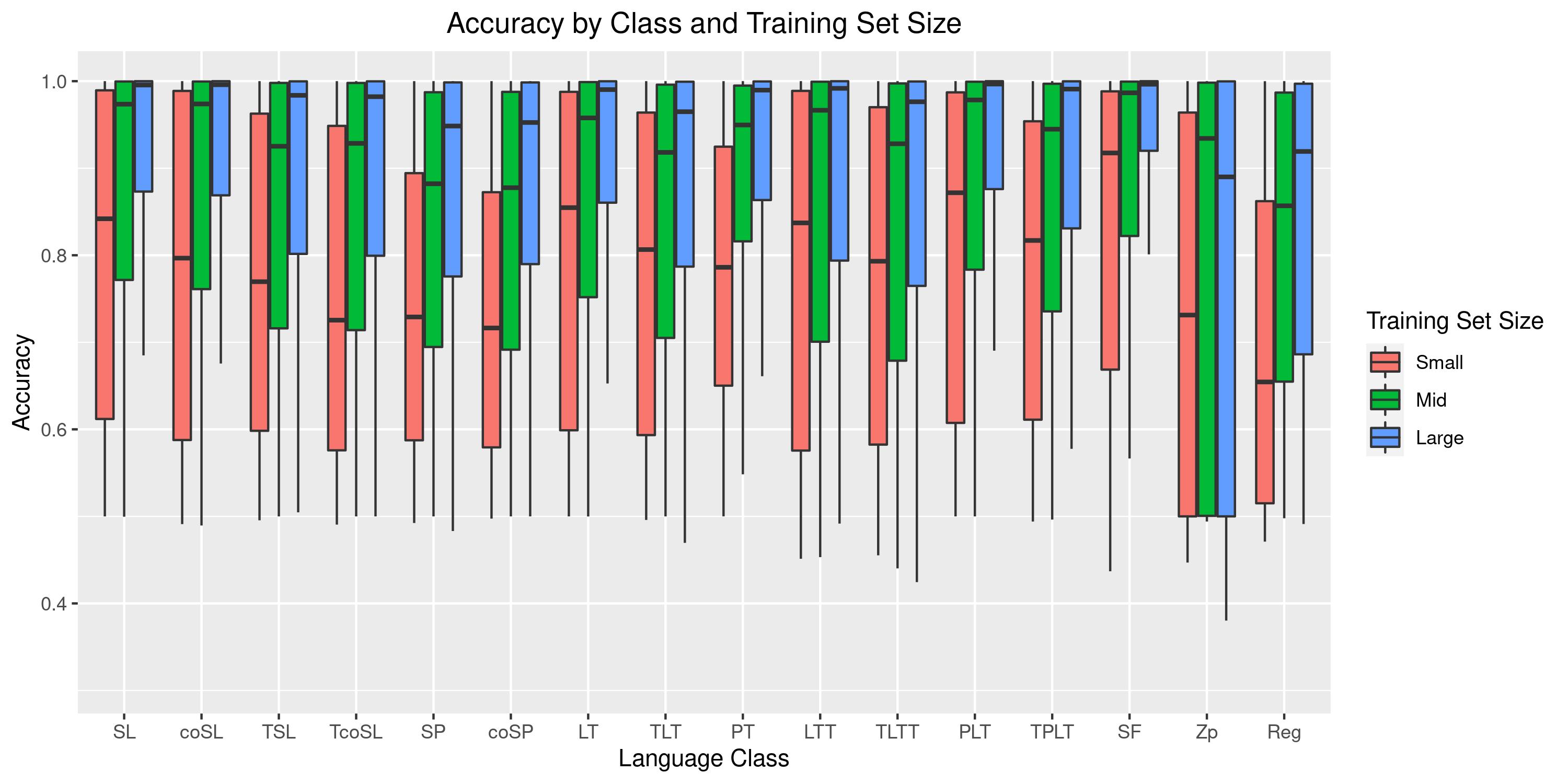}
  \caption{Accuracy by \TrainSize{} and \Class{}.}
  \label{fig:trainclass}
\end{figure}

Setting the treatment variable to \TrainSize{} and the other variables
as blocking variables, Table~\ref{tab:acc-train} shows the average
accuracy scores of each treatment level which increase as expected.
\begin{table}[ht]
  \centering
  \begin{tabular}{llll}
    \toprule
    Small (1k) & Mid (10k) & Large (100k) \\\midrule
    $0.769$    & $0.854$   & $0.887$      \\
    \bottomrule
  \end{tabular}
  \caption{Average accuracy by \TrainSize{}.}
  \label{tab:acc-train}
\end{table}
Furthermore, the Friedman test shows that the type of training set
leads to statistically significant differences in accuracy (Friedman
chi-squared = $1084.1$, df = $2$, $p$-value $<$ \num{2.2e-16}).
Post-hoc pairwise comparisons using Nemenyi-Wilcoxon-Wilcox all-pairs
test for a two-way balanced complete block design showed that each
treatment level differed significantly from the others with each
\(p\)-value less than $<$ \num{2e-16}. Visual inspection of
Figure~\ref{fig:trainclass} indicates that these results appear to
hold across individual language classes, also as expected.

\subsubsection{Complement Classes} We were also interested in
comparing the performance on the pairs of language classes SL/coSL,
TSL/TcoSL, and SP/coSP.  Recall that the languages in these classes
are paired, in the sense that for every language $L$ in class X $\in$
\{SL, SP, TSL\}, the complement $\bar{L}$ of $L$ is in
coX. Furthermore, the training, development, and test sets for every
complement language $\bar{L}$ in coX was made simply by switching the
labels in the training, development, and test sets for the
corresponding language $L$ in class X. Therefore we expected no
difference in performance.

Setting the treatment variable to \Class{} and the other variables as
blocking variables, the Friedman rank sum test shows that the type of
class leads to a statistically significant difference in accuracy
(Friedman chi-squared = $375.65$, df = $15$, $p$-value $<$
\num{2.2e-16}).  However, our question at this point is whether these
differences are found between the specific pairs of classes
highlighted above.

While the Friedman test answers the question whether any treatment
levels differ, it does not tell us where or how the treatment levels
do so. We perform post-hoc analysis using the Nemenyi-Wilcoxon-Wilcox
all-pairs test for a two-way balanced complete block design to answer
these questions. Table~\ref{tab:nww-class} in the appendix shows the
$p$-values reported by the post-hoc Nemenyi-Wilcoxon-Wilcox all-pairs
test for a two-way balanced complete block design for all language
classes. Table~\ref{tab:acc-coclass} presents the mean-aggregated
accuracies for the classes of interest here. This shows that for these
pairs of classes, there is no significant difference in accuracy as
expected.
\begin{table}[ht]
  \centering
  \begin{tabular}{ccc}
    \toprule
    \multicolumn{2}{c}{Accuracy} & All-pairs test $p$-value                \\
    \midrule
    SL                           & coSL    & \multirow{2}{*}{$0.948$} \\
    $0.862$                      & $0.855$ &                          \\
    \midrule
    SP                           & coSP    & \multirow{2}{*}{$1.000$} \\
    $0.816$                      & $0.813$ &                          \\
    \midrule
    TSL                          & TcoSL   & \multirow{2}{*}{$1.000$} \\
    $0.839$                      & $0.832$ &                          \\
    \bottomrule
  \end{tabular}
  \caption{Average accuracy for SL, coSL, SP, coSP, TSL and TcoSL classes.}
  \label{tab:acc-coclass}
\end{table}

\subsubsection{Summary}

We conclude that the sanity checks above provide evidence that our
experimental setup and factorial design are sound.

\subsection{Research Questions}

This section examines the questions posed earlier in this article that
influenced the design of MLRegTest and the neural networks used in our
experiments. These questions are summarized below.
\begin{enumerate}\itemsep-0.5ex
\item What is the effect of the type of test set (SR, SA, LR, LA)?
\item What is the effect of the language class?
  \begin{enumerate}\itemsep-0.5ex
  \item What is the effect of logical level?
  \item What is the effect of order relation (successor, tier-successor, precedence)?
  \end{enumerate}
\item What is the effect of alphabet size?
\item What is the effect of neural network architecture?
\item What is the effect of the size of the automata?
\end{enumerate}
The remainder of this section analyzes these questions one by one.

\subsubsection{The Test Set}

Setting the treatment variable to \TestType{} and the other variables
as blocking variables, Table~\ref{tab:acc-test} shows the average
accuracy scores of each treatment level.
\begin{table}[ht]
  \centering
  \begin{tabular}{cccc}
    \toprule
    SR      & LR      & SA      & LA      \\
    \midrule
    $0.944$ & $0.888$ & $0.781$ & $0.734$ \\
    \bottomrule
  \end{tabular}
  \caption{Average accuracy by \TestType{}.}
  \label{tab:acc-test}
\end{table}
The Friedman rank sum test shows that the null hypothesis that all
test set types have the same accuracies should be rejected (Friedman
chi-squared = $1965.5$, df = $3$, $p$-value $<$ \num{2.2e-16}).

\begin{figure}[ht]
  \centering
  \includegraphics[width=1\textwidth]{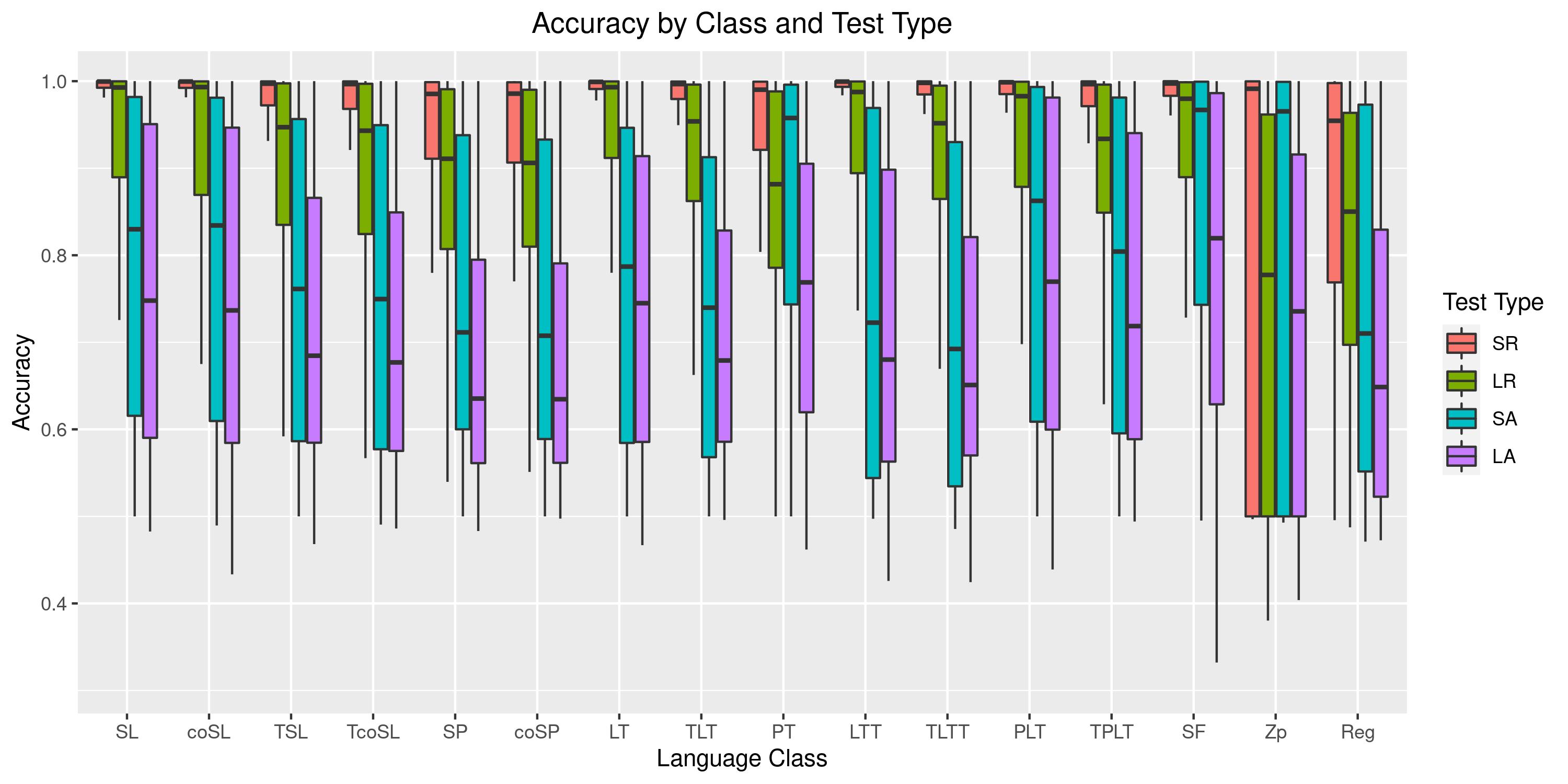}
  \caption{Accuracy by \Class{} and \TestType{}.}
  \label{fig:acc-class-test}
\end{figure}

Post-hoc pairwise comparisons using Nemenyi-Wilcoxon-Wilcox all-pairs
test for a two-way balanced complete block design showed that each
pair of treatment levels differed significantly with a $p$-value of
\num{3.6e-14} or less. It is striking that the adversariality of the
test data reduces the accuracy by approximately 15 points from the
corresponding random data, whereas the longer test data reduce
accuracy from the corresponding short data by only approximately 5
points. These differences are generally visible across the language classes as
shown in Figure~\ref{fig:acc-class-test}. The only exceptions are PT and Zp, whose mean accuracies follow the order LA < LR < SA < SR.

Table~\ref{tab:acc-test-nn} shows these trends obtain for each NN type
as well. However, it is interesting to observe the trends are more or
less pronounced depending on the NN type and the test type. The GRUs,
for example, appear overall less affected by the lengths of the test
strings than the other network types.
\begin{table}[ht]
  \centering
  \begin{tabular}[ht]{lllll}
    \toprule
                & SR    & LR    & SA    & LA    \\
    \midrule
    RNN         & 0.948 & 0.850 & 0.714 & 0.662 \\
    GRU         & 0.976 & 0.966 & 0.845 & 0.846 \\
    LSTM        & 0.947 & 0.911 & 0.748 & 0.713 \\
    Transformer & 0.961 & 0.881 & 0.758 & 0.690 \\
    \bottomrule
  \end{tabular}
  \caption{Accuracy by \TestType{} and \NNType{}.}
  \label{tab:acc-test-nn}
\end{table}
It is also interesting to observe that the GRUs outperformed the other
network types across all types of test sets.

\subsubsection{The Language Class}
\label{sec:res:lgclass}

It was already mentioned in \S\ref{sec:sanity} that the Friedman rank
sum test shows that generally the type of language class does lead to
a statistically significant difference in accuracy (Friedman
chi-squared = $831.03$, df = $15$, $p$-value $<$ \num{2.2e-16}). There
we also discussed that the Nemenyi-Wilcoxon-Wilcox all-pairs test
revealed no significant differences between the pairs SL/coSL,
SP/coSP, and TSL/TcoSL.

Table~\ref{tab:acc-class} shows mean accuracy, in decreasing order, of
each \Class{} aggregated over all experiments.
\begin{table}[ht]
  \centering
  \begin{tabular}{ll}
    \toprule
    Accuracy  & Class \\
    \midrule
    \(0.884\) & SF    \\
    \(0.866\) & PLT   \\
    \(0.862\) & PT    \\
    \(0.862\) & SL    \\
    \bottomrule
  \end{tabular}
  \begin{tabular}{ll}
    \toprule
    Accuracy  & Class \\
    \midrule
    \(0.855\) & LT    \\
    \(0.855\) & coSL  \\
    \(0.847\) & TPLT  \\
    \(0.842\) & LTT   \\
    \bottomrule
  \end{tabular}
  \begin{tabular}{ll}
    \toprule
    Accuracy  & Class \\
    \midrule
    \(0.839\) & TSL   \\
    \(0.834\) & TLT   \\
    \(0.832\) & TcoSL \\
    \(0.829\) & TLTT  \\
    \bottomrule
  \end{tabular}
  \begin{tabular}{ll}
    \toprule
    Accuracy  & Class \\
    \midrule    
    \(0.816\) & SP    \\
    \(0.813\) & coSP  \\
    \(0.781\) & Reg   \\
    \(0.770\) & Zp    \\
    \bottomrule
  \end{tabular}

  \caption{Mean accuracy in decreasing order by language \Class{}.}
  \label{tab:acc-class}
\end{table}
Table~\ref{tab:nww-class}, in the appendix, presents post-hoc pairwise
comparisons using Nemenyi-Wilcoxon-Wilcox all-pairs test showed that
many, but not all language classes differed significantly. While the
MSO-definable classes Reg and Zp are at the bottom of the list in
Table~\ref{tab:acc-class}, it is difficult to ascertain any logic
behind the order in the rest of the list.

In particular, it is striking that accuracies for languages in the SF
class are higher than any other class, even though SF is relatively
high in the complexity scale presented in Figure~\ref{fig:classes}.
This indicates that the set of languages chosen to represent SF may be
anomalously ``easy to learn.'' The TLTT and SP classes, for example,
are also star-free languages, but their overall accuracy scores are
lower than the accuracies for SF languages which are not TLTT nor SP.

In the next section we show to what extent more general properties of
the classes -- in particular, their logical level and the
model-theoretic treatment of order -- provide insight.

\subsubsection{Properties of Language Classes}

Next, we focus on whether the parameters by which we classified our
language classes (cf. Figure~\ref{fig:classes}) can help explain why
the Friedman test rejects the null hypothesis that accuracies for
\Class{} would be the same. Specifically, we investigate the impacts
of the kind of logic needed (CNL, DPL, Propositional, FO, MSO) and the
kind of representational primitive (successor, precedence,
tier-successor). We examine these properties in the aggregate as well
as for each individual neural network type.

First, we investigate the logical level. Table~\ref{tab:grp-logic}
shows how the language classes are grouped into logical levels.
\begin{table}[ht]
  \centering
  \begin{tabular}{ll}
    \toprule
    Group & Classes                \\
    \midrule
    CNL   & SL, SP, TSL            \\
    DPL   & coSL, coSP, TcoSL      \\
    PROP  & LT, PLT, PT, TLT, TPLT \\
    FO    & LTT, TLTT, SF          \\
    MSO   & Zp, Reg                \\
    \bottomrule
  \end{tabular}
  \caption{Language Classes Grouped by Logical Level}
  \label{tab:grp-logic}
\end{table}
Does accuracy generally decrease as expressivity increases logically?
If so, we would expect the accuracies to follow the
ordering CNL $\sim$ DPL $>$ Prop $>$ FO $>$ MSO.
The Friedman rank sum test shows that the logical level
leads to a statistically significant difference in accuracy (Friedman
chi-squared = $69.772$, df = $4$, $p$-value $=$ \num{2.536e-14}).
\begin{table}[ht]
  \centering
  \begin{tabular}{lllll}
    \toprule
    PROP    & CNL           & FO            & DPL           & MSO             \\
    \midrule
    $0.850$ & $0.838$       & $0.837$       & $0.832$       & $0.775$         \\
    \bottomrule
  \end{tabular}
\caption{Average accuracy by logical level in decreasing order.}
\label{tab:acc-logic}
\end{table}
Table~\ref{tab:nww-logic} shows the $p$-values from the
Nemenyi-Wilcoxon-Wilcox all-pairs test for the groups organized by
logical level.
\begin{table}[ht]
  \centering
  \begin{tabular}{lllll}
    \toprule
            & CNL           & DPL           & FO            & PROP            \\
    \midrule
    DPL     & \(0.593\)     & \centerdash   & \centerdash   & \centerdash     \\
    FO      & \(0.912\)     & \(0.141\)     & \centerdash   & \centerdash     \\
    PROP    & \num{5.5e-4} & \num{4.3e-7} & $0.013$       & \centerdash     \\
    MSO     & \num{6.4e-4} & \(0.083\)     & \num{1.3e-5} & \num{7.7e-14} \\
    \bottomrule
  \end{tabular}
  \caption{$p$-values from the Nemenyi-Wilcoxon-Wilcox all-pairs test
    for classes grouped by logical level.}
  \label{tab:nww-logic}
\end{table}
The only statistically significant differences are between PROP and
CNL, MSO and CNL, PROP and DPL, MSO and FO, and MSO and PROP. These
post-hoc comparisons indicate that the MSO level, which includes the
classes Reg and Zp, is significantly more difficult than everything
else, with the exception of DPL. Nonetheless, examination of
Tables~\ref{tab:acc-logic} shows lower levels of accuracy for classes
in the MSO group than for classes in the DPL group.

When this statistical analysis is localized to each individual
network, it reveals distinctions among them. The Friedman chi-squared
test reaches significance for the the RNN, GRU, and LSTM (p-values
equal \num{3.81e-9}, \num{3.27e-3}, \num{2.24e-4}, respectively)
but not for the Transformer (p-value = 0.023). For each network type,
mean accuracy is lowest for the REG group and highest for the PROP
group.

We conclude that the distinction made by MSO-level expressivity is
significant, but not distinctions at the lower logical levels FO,
PROP, CNL, and DPL. We are surprised there is little difference in
mean accuracy between the CNL, FO, and DPL groups and that mean
accuracy for the CNL and DPL groups was lower than mean accuracy for
the PROP group. This is observed in the individual networks
themselves, as well as altogether (which was reported in
Table~\ref{tab:acc-logic}).

Next we fix the logical level and examine the effect of the successor,
precedence, and tier-successor. Table~\ref{tab:grp-order} shows how
the language classes are grouped by the order relations.
\begin{table}[ht]
  \centering
  \begin{tabular}{ll}
    \toprule
    Group & Classes               \\
    \midrule
    SUCC  & SL, coSL, LT, LTT     \\
    PREC  & coSP, PT, SF, SP      \\
    TSUCC & TcoSL, TLT, TLTT, TSL \\
    OTHER & PLT, TPLT, Reg, Zp    \\
    \bottomrule
  \end{tabular}
  \caption{Language Classes Grouped by Order Relation}
  \label{tab:grp-order}
\end{table}
The Friedman rank sum test shows that the order relation leads to a
statistically significant difference in accuracy (Friedman chi-squared
= $28.675$, df = $3$, $p$-value $=$ \num{2.621e-6}).
\begin{table}[ht]
  \centering
  \begin{tabular}{llll}
    \toprule
    SUCC    & PREC    & OTHER   & TSUCC   \\\midrule
    $0.851$ & $0.836$ & $0.835$ & $0.833$ \\\bottomrule
  \end{tabular}
  \caption{Average accuracy for classes grouped by order relation in
    descending order.}
  \label{tab:acc-order}
\end{table}
Table~\ref{tab:nww-order} shows the $p$-values from the
Nemenyi-Wilcoxon-Wilcox all-pairs test for the groups defined with
order relations.
\begin{table}[ht]
  \centering
  \begin{tabular}{llll}
  \toprule
          & OTHER         & PREC          & SUCC          \\
    \midrule
    PREC  & \(1.000\)     & \centerdash   & \centerdash   \\
    SUCC  & \num{2.3e-4} & \num{2.8e-4} & \centerdash   \\
    TSUCC & \(0.903\)     & \(0.885\)     & \num{9.8e-6} \\
  \bottomrule
\end{tabular}
\caption{$p$-values from the Nemenyi-Wilcoxon-Wilcox all-pairs test for
   classes grouped by order relation.}
\label{tab:nww-order}
\end{table}
The SUCC group shows statistically significant differences in accuracy
as compared to the other groups.

Interestingly, when this statistical analysis is localized to
individual networks, the Friedman chi-squared test is not rejected for
any network type. This indicates that only when their results are
aggregated together can the effect of the order relation be detected.
We conclude these results indicate that patterns based on substring
(SUCC) are easier to learn than ones based on tier-substring (TSUCC),
subsequence (PREC) or some combination thereof (OTHER). However, the
effect may not be particularly strong because it was not visible when
the analysis was localized to individual networks.

\subsubsection{Alphabet size}

We also study the effect of the alphabet size. Average accuracy by
alphabet size is depicted in Table~\ref{tab:acc-alph}. The Friedman
rank sum test shows that the alphabet size leads to a
statistically significant difference in accuracy (Friedman chi-squared
= $267.82$, df = $2$, $p$-value $<$ \num{2.2e-16}).
\begin{table}[ht]
  \centering
  \begin{tabular}{ccc}
    \toprule
    64      & 16            & 4             \\
    \midrule
    $0.856$ & $0.842$       & $0.812$       \\
    \bottomrule
  \end{tabular}
  \caption{Average accuracy by alphabet size in descending order.}
  \label{tab:acc-alph}
\end{table}
Table~\ref{tab:nww-alph} shows the $p$-values from the
Nemenyi-Wilcoxon-Wilcox all-pairs test, all of which meet the standard
for significance.
\begin{table}[ht]
  \centering
  \begin{tabular}{lll}
    \toprule
            & 4             & 16            \\
    \midrule
    16      & \num{2.8e-4} & \centerdash   \\
    64      & \num{2.6e-14} & \num{4.4e-9} \\
    \bottomrule
  \end{tabular}
  \caption{$p$-values from the Nemenyi-Wilcoxon-Wilcox all-pairs test
    for alphabet size.}
  \label{tab:nww-alph}
\end{table}
There are statistically significant differences between the accuracies
on languages with the largest alphabet size as compared to the smaller
ones.

When this statistical analysis is localized to each individual
network, the Friedman chi-squared test reaches significance for the
the RNN, LSTM, and Transformer (p-values \(=\) \num{1.85e-7},
\(=\) \num{6.81e-14}, and \(<\)\ num{2.2e-16}, respectively) but not for
the GRU (p-value = 0.0158). In addition, while the levels of accuracy
of the RNN, LSTM, and Transformer models increased with alphabet size,
the levels of accuracy of the GRU did not. For the GRU, the accuracy
for the group with size 16 alphabet was almost 2 points higher than
the groups with size 4 and 64 alphabets.

We conclude that in general patterns become easier to learn the larger
the alphabet, there are network models that can subvert this trend.

\subsubsection{What is the effect of the neural network?}

The Friedman rank sum test shows that the neural network type also
leads to a statistically significant difference in accuracy (Friedman
chi-squared = $880.2$, df = $3$, $p$-value $<$ \num{2.2e-16}). Those
accuracies are shown in Table~\ref{tab:acc-network}.
\begin{table}[ht]
  \centering
  \begin{tabular}{llll}
    \toprule
    GRU     & Transformer & LSTM    & RNN \\
    \midrule
    $0.906$ & $0.826$     & $0.825$ & $0.790$    \\
    \bottomrule
  \end{tabular}
  \caption{Accuracies by neural network type in descending order.}
    \label{tab:acc-network}
  \end{table}

  Table~\ref{tab:nww-network} shows the $p$-values from the
Nemenyi-Wilcoxon-Wilcox all-pairs test.
\begin{table}[ht]
  \centering
  \begin{tabular}{llll}
    \toprule
                & RNN               & GRU              & LSTM        \\
    \midrule
    GRU         & \(<\)\num{2e-16} & \centerdash      & \centerdash \\
    LSTM        & \num{3.7e-14}     & \(<\)\num{2e-16} & \centerdash \\
    Transformer & \(<\)\num{2e-16}  & \(<\)\num{2e-16} & \(0.130\)   \\
    \bottomrule
  \end{tabular}
  \caption{$p$-values from the Nemenyi-Wilcoxon-Wilcox all-pairs test
    for neural network type.}
  \label{tab:nww-network}
\end{table}
All pairwise comparisons are significantly different except for the
one between the LSTM and the Transformer.

Since the above results aggregate over all training set sizes, we
repeated the above analysis by partitioning the data according to the
training set size. After restricting to different training set sizes,
the Friedman rank sum test continued to show that the network type
significantly impacted accuracy with $p<$\num{2.2e-16}. The accuracies
are shown in Table~\ref{tab:acc-network-by-train}.
\begin{table}[ht]
  \centering
  \begin{tabular}{llll}
    \toprule
                & Small          & Mid            & Large          \\
    \midrule
    Simple RNN  & \(0.736\)      & \(0.796\)      & \(0.839\)      \\
    GRU         & \textbf{0.843} & \textbf{0.934} & \textbf{0.939} \\
    LSTM        & \(0.720\)      & \(0.855\)      & \(0.901\)      \\
    Transformer & \(0.779\)      & \(0.830\)      & \(0.867\)      \\
    \bottomrule
  \end{tabular}
  \caption{Average accuracy by neural network type and training
    size. Bold-faced scores are the highest in each column, and are
    statistically significantly different than non-bold scores.}
  \label{tab:acc-network-by-train}
\end{table}
This analysis reveals that GRUs outperform the other networks on all
training regimes.

\begin{figure}[ht]
  \centering
  \includegraphics[width=1\textwidth]{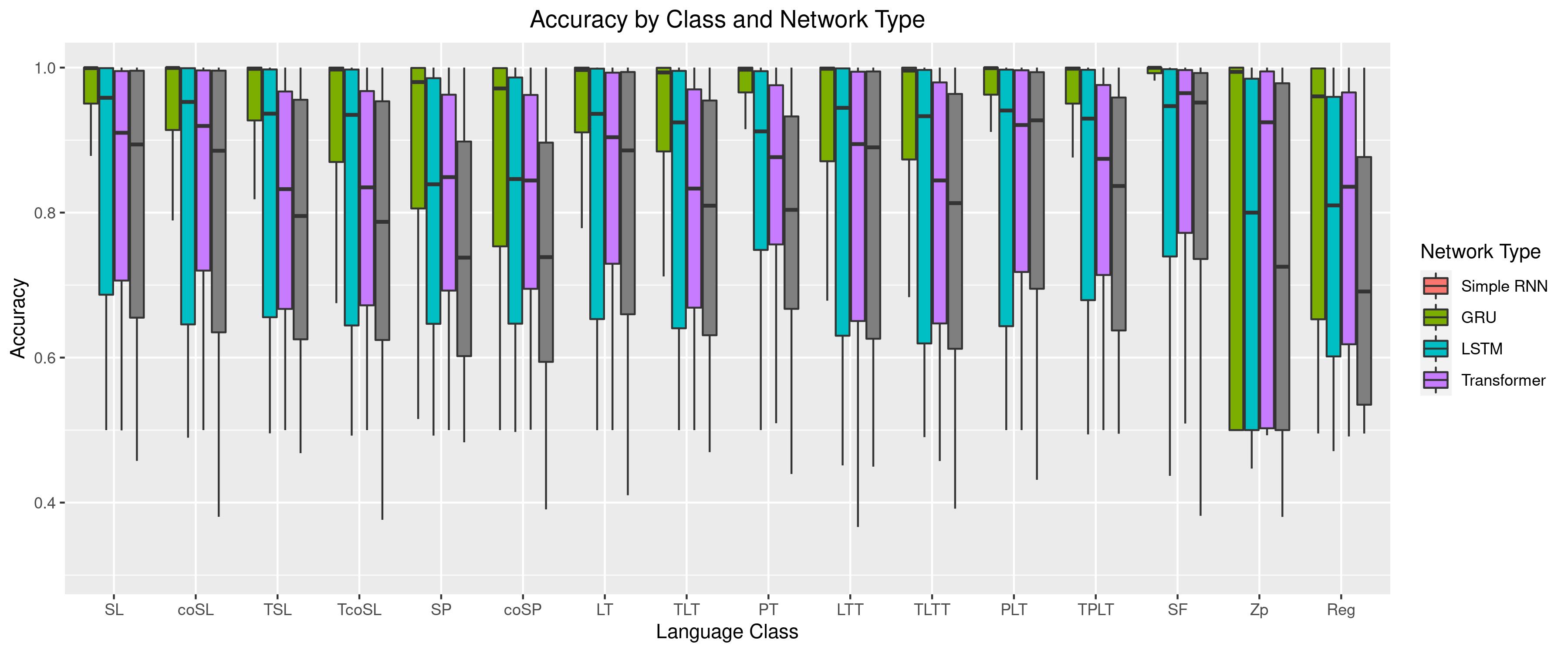}
  \caption{Accuracy by Class and Neural Network}
  \label{fig:acc-class-nn}
\end{figure}

\begin{figure}[ht]
  \centering
  \includegraphics[width=1\textwidth]{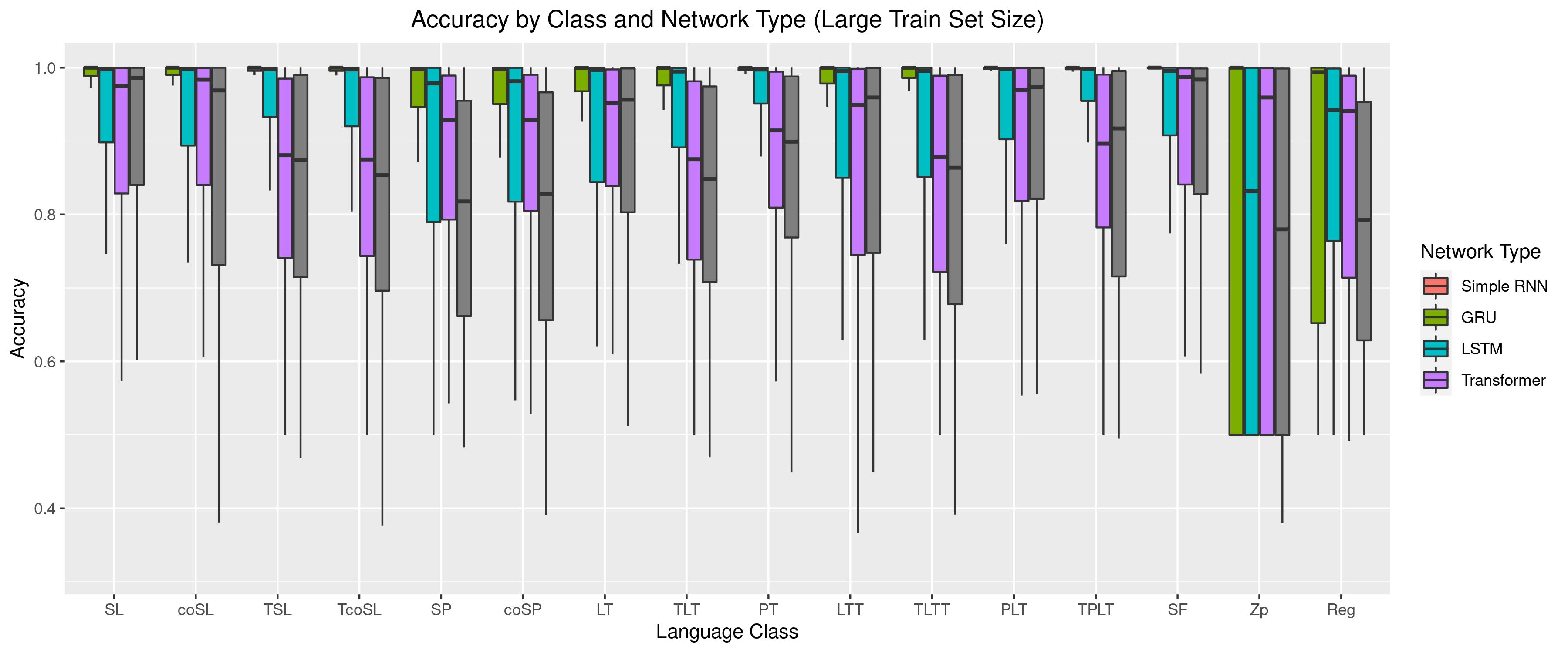}
  \caption{Accuracy by Class and Neural Network on the Large Training
    Set.}
  \label{fig:acc-class-nn-large}
\end{figure}

Figures~\ref{fig:acc-class-nn} and~\ref{fig:acc-class-nn-large}
provide a visualization of the performance of the neural networks on
each language class aggregating across training regimes and for the
Large training set, respectively.
Visual inspection reveals that there is significant variation both
across language classes, and across network types within language
classes. Nonetheless, it is clear that the GRU outperforms every other
network on all classes. 
Furthermore, when trained on the Large dataset, the average accuracy
for the GRUs, across all classes, is close to 100\%, though it is
visually evident that GRU performance on the Zp and Reg classes shows
considerable more variation than on the other classes.

\subsubsection{Grammar size}

While the size of the grammar of the formal language was not a
treatment variable, we chose to examine it anyway. The number of
states of the minimal DFA is a standard measure for the size of the
grammar for a regular language. We also considered the number of
states of the syntactic monoid of the minimal
DFA. Table~\ref{tab:sizes} provides summary statistics on these size
measures of the representations of the MLRegTest languages.

We were interested in how well accuracy scores were inversely
correlated with grammar size. Table~\ref{tab:corr-size} shows these
overall correlations calculated using Spearman's rank correlation for
all the training sets, as well as for the Large training set. 
\begin{table}[ht]
  \centering
  \begin{tabular}{lll}
    \toprule
                                & All train & Large train \\
    \midrule
    DFA size $\sim$ accuracy    & $-0.104$  & $-0.173$    \\
    monoid size $\sim$ accuracy & $-0.098$  & $-0.165$    \\
    \bottomrule
  \end{tabular}
  \caption{Correlations between accuracies and the size of the
    target pattern measured by the size of the minimal DFA and its
    syntactic monoid on all training sets and the large training
    set.}
  \label{tab:corr-size}
\end{table}
These results revealed a statistically significant inverse correlation
(\(p<\)\num{2e-16}). In other words, it is the case that generally
accuracy decreases the larger the automata. However, these
correlations are much closer to 0 than to 1, which is indicative of a
weak effect.

We calculated other correlations making finer distinctions by network
type, test type, and training size. The strongest correlation we found
was for the GRU trained on the Large training set and evaluated on the
Short Adversarial test set. In this block of data, Spearman's rank
correlation for the minimal DFA was $-0.498$ and for its syntactic
monoid $-0.509$. These correlations are noticeably larger, and are a
good indication that automata size influences GRU performance in this
learning scenario.

\section{Discussion} 

The results of \S\ref{sec:res} support the following conclusions. 
\begin{itemize}\itemsep-0.5ex
\item Regardless of language class and neural network type, high
  performance on Random test sets does not imply correct
  generalization as measured by performance on the Adversarial test
  sets (Table~\ref{tab:acc-test}).
\item Learning classifiers which depend on counting modulo $n$ is more
  difficult for neural ML systems (Table~\ref{tab:acc-logic}).
\item Learning classifiers for languages which only need to keep track
  of substrings (i.e. logically invoking only the successor relation)
  is easier for neural ML systems than learning classifiers for
  languages needing to keep track of certain kinds of non-local
  dependencies (i.e those which logically invoke the tier-successor or
  precedence relation) (Table~\ref{tab:acc-order}).
\item Learning classifiers for languages with smaller alphabets is
  generally more difficult for neural ML systems
  (Table~\ref{tab:acc-alph}).
\item Classification ability of neural ML systems correlates weakly
  with the size of the minimal DFA and its syntactic monoid
  (Table~\ref{tab:corr-size}).
\item Across all facets of MLRegTest, the GRU is overall the best
  performing architecture (Tables~\ref{tab:acc-network} -
  \ref{tab:acc-network-by-train}). That the GRUs are also the models
  with the fewest parameters (Table~\ref{params-table}) makes this an
  especially notable result.
\end{itemize}

These results provide evidence that MLRegTest is a valuable benchmark
for ML systems being used for sequence classification. While the GRUs
overall performed the best, the analysis here shows there is
considerable room for improvement. In addition to improving
generalization ability as measured by the adversarial test sets,
MLRegTest has helped identify some of the more challenging sequential
patterns. These include those which count modulo \(n\), those with
other kinds of non-local dependencies, those with smaller alphabets,
and those represented by larger DFA.

Some of these results may find an account in terms of related
research. The adversarial test sets can be thought of as demanding a
higher degree of sensitivity in the sense of
\citet{HahnJurafskyFutrell2021} and
\citet{bhattamishra-etal-2023-simplicity}. Similarly, research on the
expressivity of network architectures has shown that certain kinds of
transformers cannot express patterns that count modulo \(n\)
\citep{yang+2024} (see also \citet{MerrillSabharwal2023} and
\citet{Strobl+2024a}).

Regarding the difficulty of learning patterns which count modulo
\(n\), our results are both consistent with, and inconsistent with,
earlier research. One the one hand, these results are in line with
those reported by \citet{bhattamishra-etal-2020-ability}, who found
transformers struggle with these languages. On the other hand, these
results contrast with \citeauthor{Deletang+2022}'s
\citeyear{Deletang+2022} results, which found RNNs capable of learning
such patterns. Their experiments differed from ours in both the
training and testing data. For instance, in their experiments, NNs
were exposed to training data of strings of length less than
19. Shorter sequences may be especially important in training; this is
a topic to be studied more carefully in future research.

Another question for future research is determining the relative
impacts of properties of a pattern in addressing its learning
difficulty. In this regard, it is interesting to note that some of the
aforementioned factors conflict. For example, counting modulo two only
requires a DFA with 2 states (Figure~\ref{fig:even-a}). Nonetheless,
despite its small size, counting module \(n\) was also shown to be a
generally more difficult pattern for neural ML systems to learn.

Finally, the fact that the GRUs outperformed the other neural networks
while having fewer parameters by the thousands, or in some cases, by
hundreds of thousands, indicates that addressing these challenges need
not come in the form of feeding ever bigger models with more data. To
drive this point home, consider that on the language which recognizes
strings with an even number of \(a\)s with an alphabet of size 16
(language 16.16.Zp.2.1.0), when presented with the Small training
regime and tested on the LA test set, the GRU obtained 99.96\%
accuracy, the transformer 80.92\% accuracy, and the LSTM and RNN were
at chance. When one considers the number of trainable parameters these
models have (Table~\ref{params-table}), it seems clear that \emph{how}
the parameters interact to make predictions can be much more important
than \emph{how many} parameters there are.

\section{Conclusion}
This article presented a new benchmark for machine learning systems on
sequence classification. This benchmark, called MLRegTest, contains
training, development, and test sets from 1,800 regular languages
spread across 16 subregular classes. These languages are organized
according to their logical complexity (monadic second order, first
order, propositional, or propositional under additional restrictions)
and the kind of logical
literals (string, tier-string, subsequence, or combinations
thereof). The logical complexity and choice of literal provides a
systematic way to understand different kinds of long-distance
dependencies in regular languages, and therefore to understand the
capacities of different ML systems to learn such long-distance
dependencies.

In addition to providing three nested training sets for each language,
MLRegTest provides four test sets according to two binary parameters:
string length (short/long) and data generation (random/adversarial).

Finally, we examined the performance of different neural networks
(simple RNNs, LSTMs, GRUs, Transformers) on MLRegTest. While there is
much variation in the performance, some statistical trends were
clear. First, the neural networks generally performed worse on the
adversarial test sets; these contained pairs of strings of string edit
distance 1 with the property that one belonged to the target language
and the other did not. These results imply the networks did not
generalize correctly despite very high accuracies on the random test
sets.

Second, GRUs generally outperformed the other network models across
all languages, training regimes, and test sets. The fact that these
networks did not possess many parameters, relative to the LSTMs and
Transformer networks, indicates that improved performance does not
require larger networks. Nonetheless, even the GRU network has room
for improvement on MLRegTest.

Third, the formal properties of the languages themselves were
important in determining their learning difficulty. It was shown that
neural networks have difficulty learning periodic regular languages;
i.e those that require monadic second order logic. Another conclusion
was that the neural networks generally performed worse on classifying
strings on languages defined in terms of the successor relation as
opposed to other relations representing order in strings.  The number
of symbols in the alphabet was also shown to make a
difference. Neither the size of the minimal DFA nor the size of its
syntactic monoid correlated well with NN performance, though a weak
correlation was detected. Future research and controlled experiments
are needed to better tease apart these factors.

The overall results also raise questions in formal lanugage theory.
Recall that the PLT class,
which uses propositional logic with both successor and precedence,
subsumes the LTT class,
which uses first-order logic but with successor alone.
In other words, a language cannot be associated with a logical level
independent from the ordering relations used to describe it.
Given that the first-order patterns in MLRegTest
appear easier to learn than expected,
one may wonder whether there is some yet-unknown class
which uses only propositional logic but contains these patterns,
or even subsumes our first-order classes.
The issue also arises
with respect to sampling data from a formal language.
There are several grammars compatible with any finite data set.
This leads to a variety of questions.
What is the degree of overlap,
i.e.\@ how well can a simpler logic approximate a more complex one?
To what extent is accuracy affected by the grammar
used to generate samples?

Altogether, we hope that MLRegTest provides a useful tool for
researchers in machine learning interested in sequence
classification. We believe that an ML system which can achieve near
perfect accuracy on all test sets for all languages with only the
smaller training regimes will be revolutionary.

\acks{Language class verification, data set creation, and neural
  network training and evaluation were completed on the Stony Brook
  SeaWulf HPC cluster maintained by Research Computing and
  Cyberinfrastructure, and the Institute for Advanced Computational
  Science at Stony Brook University and made possible by NSF grant
  \#1531492.

  We also thank R\'emi Eyraud, Guillaume Rabusseau, and the audiences
  at LearnAut 2022, the FLaNN group, the All Things Language and
  Computation seminar at Stony Brook University, the Mathematical
  Linguistics Recreation Group at Stony Brook University, and the
  Hubert Curien Laboratory at Jean Monnet University for their
  valuable advice and feedback. We are also grateful to the associate
  editor and three anonymous reviewers whose expert advice greatly
  improved this research.}


\newpage

\appendix
\section*{Appendix}\label{sec:appendix}


\subsection*{Languages used in Hyperparameter Search}

\begin{table}[ht]
  \centering
  \begin{tabular}{l}
  \toprule
    16.07.TLT.4.1.3 \\
    16.07.TLT.4.1.6 \\
    16.07.TLTT.4.2.3 \\
    16.07.TLTT.4.2.6 \\
    16.07.TPLT.4.2.3 \\
    16.07.TPLT.4.2.6 \\
    16.07.TSL.4.1.3 \\
    16.07.TSL.4.1.6 \\
    16.07.TcoSL.4.1.3 \\
    16.07.TcoSL.4.1.6 \\
    16.16.LT.4.1.3 \\
    16.16.LT.4.1.6 \\
    16.16.LTT.4.2.3 \\
    16.16.LTT.4.2.6 \\
    16.16.PLT.4.2.3 \\
    16.16.PLT.4.2.6 \\
    16.16.PT.4.1.3 \\
    16.16.PT.4.1.6 \\
    16.16.Reg.0.0.3 \\
    16.16.Reg.0.0.6 \\
    16.16.SF.0.0.3 \\
    16.16.SF.0.0.6 \\
    16.16.SL.4.1.3 \\
    16.16.SL.4.1.6 \\
    16.16.coSL.4.1.3 \\
    16.16.coSL.4.1.6 \\
    16.16.SP.4.1.3 \\
    16.16.SP.4.1.6 \\
    16.16.coSP.4.1.3 \\
    16.16.coSP.4.1.6 \\
    16.16.Zp.3.1.3 \\
    16.16.Zp.3.1.6 \\
    \bottomrule
  \end{tabular}
\caption{Languages used in the hyperparameter search.}
\label{tab:search-lgs}
\end{table}

\subsection*{Statistical Tables}

Table~\ref{tab:nww-class} lists statistical results corresponding with the analyses of \Cref{sec:res}.
\begin{table}[ht]
  \centering
  \begin{tabular}{lllllllll}
    \toprule
          & coSL          & coSP          & LT            & LTT           & PLT           & PT            & Reg              & SF            \\
    \midrule
    coSP  & \num{3.6e-7} & \centerdash   & \centerdash   & \centerdash   & \centerdash   & \centerdash   & \centerdash      & \centerdash   \\
    LT    & 1.000         & \num{6.2e-8} & \centerdash   & \centerdash   & \centerdash   & \centerdash   & \centerdash      & \centerdash   \\
    LTT   & 0.801         & 0.010         & 0.603         & \centerdash   & \centerdash   & \centerdash   & \centerdash      & \centerdash   \\
    PLT   & 0.786         & \num{3.7e-13} & 0.923         & 0.004         & \centerdash   & \centerdash   & \centerdash      & \centerdash   \\
    PT    & 0.979         & 0.001         & 0.910         & 1.000         & 0.028         & \centerdash   & \centerdash      & \centerdash   \\
    Reg   & \num{2.1e-13} & 0.124         & \num{1.6e-13} & \num{3.0e-10} & \num{6.0e-14} & \num{7.0e-12} & \centerdash      & \centerdash   \\
    SF    & 0.005         & \num{1.3e-13} & 0.015         & \num{1.3e-7} & 0.830         & \num{2.6e-6} & \(<\)\num{2e-16} & \centerdash   \\
    SL    & 0.948         & \num{3.0e-12} & 0.991         & 0.017         & 1.000         & 0.091         & \num{1.5e-13}    & 0.575         \\
    SP    & \num{2.8e-6} & 1.000         & \num{5.3e-7} & 0.034         & \num{2.5e-12} & 0.005         & 0.044            & \num{1.2e-13} \\
    TcoSL & \num{4.5e-5} & 1.000         & \num{9.8e-6} & 0.155         & \num{1.1e-10} & 0.034         & 0.007            & \num{2.0e-13} \\
    TLT   & \num{1.8e-4} & 0.999         & \num{4.2e-5} & 0.292         & \num{7.7e-10} & 0.081         & 0.002            & \num{1.2e-13} \\
    TLTT  & \num{3.2e-6} & 1.000         & \num{6.1e-7} & 0.037         & \num{3.0e-12} & 0.006         & 0.041            & \num{1.2e-13} \\
    TPLT  & 0.910         & 0.004         & 0.763         & 1.000         & 0.010         & 1.000         & \num{5.9e-11}    & \num{4.9e-7} \\
    TSL   & 0.006         & 0.879         & 0.002         & 0.830         & \num{1.3e-7} & 0.471         & \num{6.2e-5}    & \num{2.7e-13} \\
    Zp    & 0.002         & 0.957         & \num{5.8e-4} & 0.677         & \num{2.8e-8} & 0.308         & \num{2.0e-4}    & \num{1.8e-13} \\
    \midrule
    \midrule
          & SL            & SP            & TcoSL         & TLT           & TLTT          & TPLT          & TSL                              \\
    \midrule
    coSP  & \centerdash   & \centerdash   & \centerdash   & \centerdash   & \centerdash   & \centerdash   & \centerdash                      \\
    LT    & \centerdash   & \centerdash   & \centerdash   & \centerdash   & \centerdash   & \centerdash   & \centerdash                      \\
    LTT   & \centerdash   & \centerdash   & \centerdash   & \centerdash   & \centerdash   & \centerdash   & \centerdash                      \\
    PLT   & \centerdash   & \centerdash   & \centerdash   & \centerdash   & \centerdash   & \centerdash   & \centerdash                      \\
    PT    & \centerdash   & \centerdash   & \centerdash   & \centerdash   & \centerdash   & \centerdash   & \centerdash                      \\
    Reg   & \centerdash   & \centerdash   & \centerdash   & \centerdash   & \centerdash   & \centerdash   & \centerdash                      \\
    SF    & \centerdash   & \centerdash   & \centerdash   & \centerdash   & \centerdash   & \centerdash   & \centerdash                      \\
    SL    & \centerdash   & \centerdash   & \centerdash   & \centerdash   & \centerdash   & \centerdash   & \centerdash                      \\
    SP    & \num{4.1e-11} & \centerdash   & \centerdash   & \centerdash   & \centerdash   & \centerdash   & \centerdash                      \\
    TcoSL & \num{1.5e-9} & 1.000         & \centerdash   & \centerdash   & \centerdash   & \centerdash   & \centerdash                      \\
    TLT   & \num{9.8e-9} & 1.000         & 1.000         & \centerdash   & \centerdash   & \centerdash   & \centerdash                      \\
    TLTT  & \num{4.9e-11} & 1.000         & 1.000         & 1.000         & \centerdash   & \centerdash   & \centerdash                      \\
    TPLT  & 0.037         & 0.016         & 0.084         & 0.176         & 0.017         & \centerdash   & \centerdash                      \\
    TSL   & \num{1.2e-6} & 0.977         & 1.000         & 1.000         & 0.980         & 0.686         & \centerdash                      \\
    Zp    & \num{2.9e-7} & 0.996         & 1.000         & 1.000         & 0.996         & 0.509         & 1.000                            \\
    \bottomrule
  \end{tabular}
  \caption{P-values from the Nemenyi-Wilcoxon-Wilcox all-pairs test with treatment variable \Class{}.}
  \label{tab:nww-class}
\end{table}

\clearpage 

\vskip 0.2in
\bibliography{ref}

\end{document}